\documentclass{article}

    \usepackage[preprint, nonatbib]{neurips_2020}

\usepackage[utf8]{inputenc} 
\usepackage[T1]{fontenc}    
\usepackage{hyperref}       
\usepackage{url}            
\usepackage{booktabs}       
\usepackage{amsfonts}       
\usepackage{nicefrac}       
\usepackage{microtype}      
\usepackage{xcolor}         
\usepackage{graphicx}       
\usepackage{caption}        
\usepackage{subcaption}     
\usepackage{multirow}       
\usepackage{textcomp}
\usepackage[ruled,vlined]{algorithm2e} 
\usepackage{amsmath}
\DeclareMathOperator*{\argmax}{arg\,max}
\DeclareMathOperator*{\argmin}{arg\,min}

\SetCommentSty{mycommfont}

\newcommand{\var}{\texttt}
\newcommand{\func}{\texttt}

\title{A Study of Compositional Generalization in Neural Models}

\author{
    Tim Klinger\thanks{equal contribution.} \\
    Thomas J. Watson Research Center \\
    IBM Research AI \\
    Yorktown, NY, USA \\
    \texttt{tklinger@us.ibm.com} \\
    \And
    Dhaval Adjodah\textsuperscript{*}\\
    MIT Media Lab \\
    Cambridge, MA \\
    \texttt{dval@mit.edu} \\
    \And
    Vincent Marois \\
    Thomas J. Watson Research Center \\
    IBM Research AI \\
    Yorktown, NY, USA \\
    \texttt{vincent.marois@ibm.com} \\
    \And
    Josh Joseph \\
    MIT Media Lab \\
    Cambridge, MA \\
    \texttt{jmjoseph@mit.edu} \\
    \And
    Matthew Riemer \\
    Thomas J. Watson Research Center \\
    IBM Research AI \\
    Yorktown, NY, USA \\
    \texttt{mdriemer@us.ibm.com} \\
    \And
    Alex `Sandy' Pentland \\
    MIT Media Lab \\
    Cambridge, MA \\
    \texttt{pentland@mit.edu} \\
    \And
    Murray Campbell \\
    Thomas J. Watson Research Center \\
    IBM Research AI \\
    Yorktown, NY, USA \\
    \texttt{mcam@us.ibm.com}
}

\begin{document}

\maketitle

\begin{abstract}
Compositional and relational learning is a hallmark of human intelligence, but one which presents challenges for neural models. One difficulty in the development of such models is the lack of benchmarks with clear compositional and relational task structure on which to systematically evaluate them. In this paper, we introduce an environment called ConceptWorld, which enables the generation of images from compositional and relational concepts, defined using a logical domain specific language. We use it to  generate images for a variety of compositional structures: 2x2 squares, pentominoes, sequences, scenes involving these objects, and other more complex concepts. We perform experiments to test the ability of standard neural architectures to generalize on relations with compositional arguments as the compositional depth of those arguments increases and under substitution. We compare standard neural networks such as MLP, CNN and ResNet, as well as state-of-the-art relational networks including WReN and PrediNet in a multi-class image classification setting. For simple problems, all models generalize well to close concepts but struggle with longer compositional chains. For more complex tests involving substitutivity, all models struggle, even with short chains. In highlighting these difficulties and providing an environment for further experimentation, we hope to encourage the development of models which are able to generalize effectively in compositional, relational domains.
\end{abstract}

\section{Introduction}
Humans have a relational and compositional view which enables them to better manage the complexity of the world~\cite{trees1977forest, fodor1988connectionism, spelke2007core}, but extracting this view from images and text is a challenge for neural networks~\cite{lake2017generalization, loula2018rearranging}. Recent work has begun to address this challenge through the development of relational datasets and frameworks - for example \cite{johnson2017clevr, hudson2019gqa} for Visual Question Answering. These have in turn spurred research for novel relational models. We believe there is a similar need for datasets of compositional, relational concepts, however we know of no work to enable formal specification and image generation for them. In this paper, we describe such an environment, which we call ConceptWorld.

Concepts in ConceptWorld are specified hierarchically in a logical language. We see several benefits in this approach: (1) it makes it easier to define concepts whose structure is clear (to the author and others) because it is logical and  declarative rather than procedural (it specifies what it is, not how to compute it);  (2) it allows rapid prototyping and experimentation because it reduces the amount of code that needs to be written; (3) it supports easy sharing between domains as concepts are hierarchical and lower level ones can be re-used. As a test of ConceptWorld's ability to represent concepts of interest, we use it to recreate the key-lock task of Box-World \cite{zambaldi2018deep} in our setting.

\begin{figure}
  \centering
  \includegraphics[width=\linewidth]{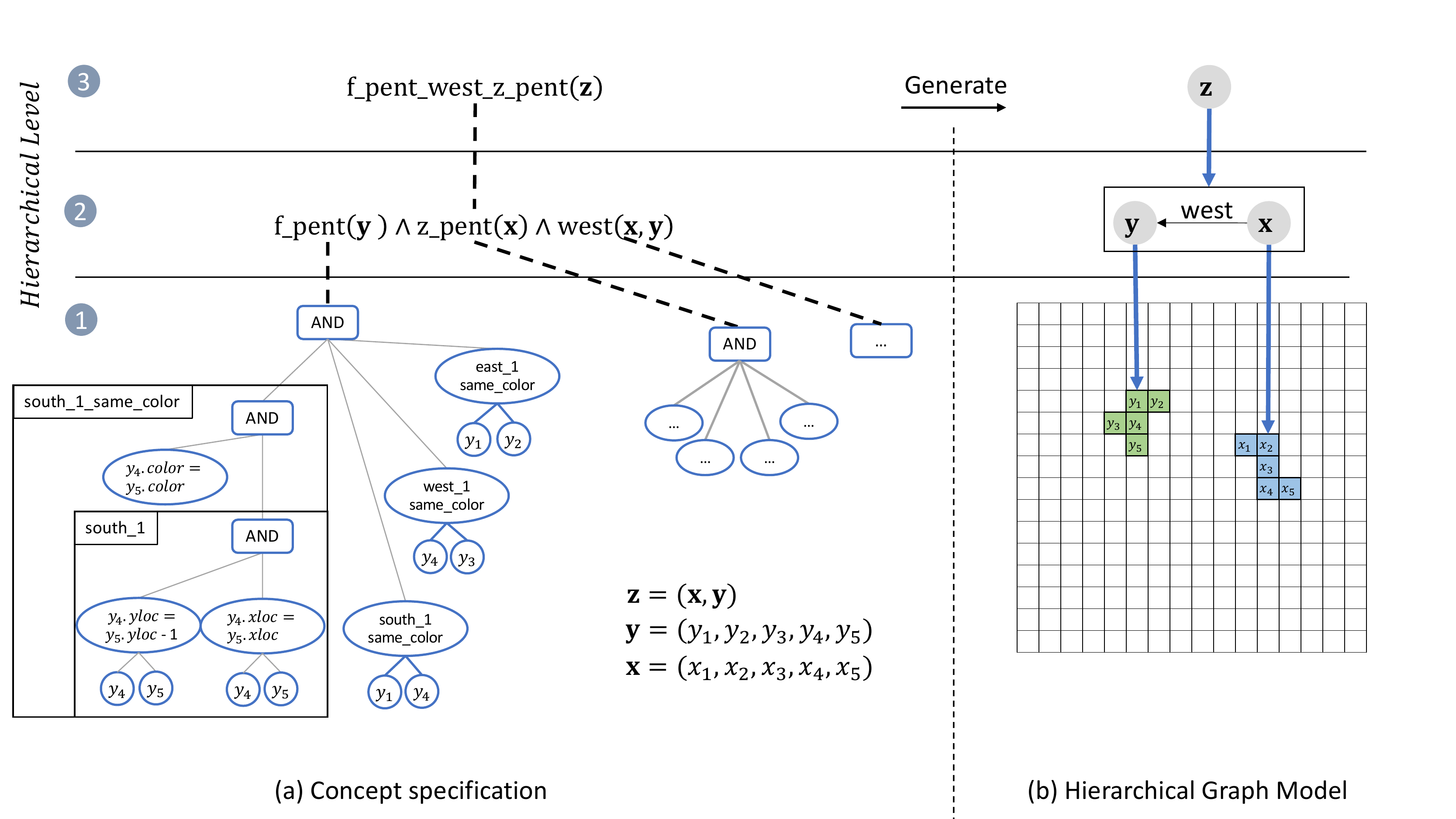}
  \caption{Representation of the $\texttt{f-pent-west-z-pent}$ concept. (a): View of the concept specification. The compositional levels are labelled 1-3. (b) A hierarchical scene graph with accompanying generated image (32 x 32 pixel).}
  \label{fig:f_pent}
\end{figure}

We consider two types of (zero-shot) compositional generalization \cite{hupkes2020compositionality}. In the  \textit{productivity} experiments, we examine compositional generalization of a concept relation to greater compositional depths than have been seen in training (for example, from a length 2 sequence of squares to a length 3 sequence).  In the \textit{substitutivity} experiments, we maintain the same depth of composition from training to test but change the object being composed.  For example, given training on (1) a concept which consists of red squares and f-pentominoes \footnote{A pentomino is a set of 5 adjacent points. Pentominoes come in different shapes which are labeled with letters (like "f") which they resemble. See Figure \ref{fig:f_pent}b.}, and (2) sequences of red squares, can the model generalize to same-length sequences of f-pentominoes? Although the concepts we use in these experiments are artificial, the compositional patterns we discuss occur frequently in natural and man-made images (the patterned fabric of a dress for example) and the ability to recognize them in a way which compositionally generalizes is one which we believe is central to more efficient learning and effective generalization.

We conduct experiments to evaluate compositional generalization on four example domains (specified using ConceptWorld) and  a variety of standard models (MLP, CNN, ResNet~\cite{he2016deep}) and newer ones designed specifically to extract relational representations (WReN~\cite{santoro2018measuring}, PrediNet~\cite{shanahan2019explicitly}). None of the models we evaluate are explicitly biased to encourage compositional generalization.

To summarize, our paper makes the following contributions:
\begin{itemize}
\item ConceptWorld: A concept specification language and generator for compositional relational concepts.
\item Four tasks each with their own domain: two which test compositional \textit{productivity} (experiment 1: pure and mixed sequences; and experiment 2: Box-World sequences); and two  which test compositional \textit{substitutivity} (experiment 3: 2x2->4x4 patterned squares; and experiment 4: sequence substitutions). 
\item An evaluation of standard and relational models on these domains. Our results demonstrate that the evaluated models struggle in these settings, including recently proposed relational models. This suggests that new research is needed to encourage compositional generalization with neural models and ConceptWorld provides a much needed test-bed in this direction.
\end{itemize}


\section{Related Work}
\label{sec:related-work}

We focus here on work related to compositionality of neural models, compositional generalization, relational learning in neural models, and reasoning tests. 


\cite{hupkes2020compositionality} discusses a variety of interpretations of the term "compositional generalization". Their taxonomy includes "productivity" (generalization to deeper compositional depths) and "substitutivity" (generalization to objects of the same type as seen in training), which we evaluate here. In \cite{lake2017generalization}, they develop a compositional dataset called SCAN, and test compositional generalization on recurrent neural network models. This is extended in \cite{loula2018rearranging}. A framework for measuring compositional generalization is proposed in \cite{santoro2018measuring} using Raven-style matrices. It is based on two principles: (1) minimize the difference between train and test distributions for "atomic" concepts (the base concepts used in compositions); and (2) maximize the difference between train and test distributions for "compound" (composite) concepts and uses a procedural language with logical operators. We have attempted to follow these principles in the construction of our experiments. Similarly, \cite{shanahan2019explicitly} introduces the Relations Game as a set of binary image classification tasks, where the aim is to label whether a specified relation holds. They use visual shapes such as pentominoes and hexominoes to test substitutivity, when the unique relation is given. We use the WReN and PrediNet models introduced in these papers as baselines in our experiments.

Compositional generalization is considered in \cite{chang2018automatically} where they introduce a model called Compositional Recursive Learner (CRL) for multi-task learning. The emphasis there is on transformations such as language translation rather than classification, but the CRL model or a Routing Network \cite{rosenbaum2017routing} might provide a good starting point for an architecture capable of generalizing more systematically than the ones considered in our experiments. Unfortunately, there are several known challenges for stabilizing the learning of these models \cite{challenges}. 

In relational reasoning, the Visual Question Answering (VQA) ~\cite{antol2015vqa} setting is used frequently. \cite{johnson2017clevr} introduces the CLEVR dataset, linking templatized natural language questions to elementary visual scenes. Critics of such artificial language datasets \cite{manjunatha2019explicit, zhang2016yin, jabri2016revisiting}  have pointed to the linguistic and semantic simplicity of the questions, as well as tendencies in the answers distributions as circumventing the need for "true" visual reasoning. \cite{hudson2019gqa} introduced the GQA dataset, to remediate some of these shortcomings.

VQA research has spawned the development of several relevant models. In \cite{hudson2018compositional}, they show that iterative, attention-based reasoning leads to more data-efficient learning. \cite{hudson2019learning, mao2019neuro} draw on the strengths of neural and symbolic AI to perform visual reasoning. \cite{santoro2017simple} proposes a simple neural module to reason about entities (visual or textual) and their relations. The Neuro-Symbolic Concept Learner (NSCL) \cite{mao2019neuro} is a multi-modal model which can learn visual concepts from training on images paired with textual questions and answers. In \cite{higgins2017scan}, they propose a generative model called SCAN (unrelated to the dataset SCAN), based on a $\beta$-VAE which can learn grounded hierarchical concepts from a small number symbol-image pairs.

\section{ConceptWorld}
\label{sec:conceptworld}

\subsection{Definitions}
A \emph{concept} is a unary or binary relation over objects. Objects can be simple points or vectors of objects, themselves possibly vectors. Concepts whose parameters are points are called \emph{primitive} concepts; while those with parameters which are vectors satisfying lower-level concepts are called \emph{higher-order}. For example, the unary relation $\texttt{red}(x_1)$ (The object $x_1$ is red) and the binary relation $\texttt{west}(x_1,x_2)$ ($x_2$ is west of $x_1$), for $x_1$ and $x_2$ grid points are both primitive concepts. See Table \ref{tab:some-concepts} for an example of a higher order unary concept whose argument is a vector $\mathbf{x} = (x_1,x_2,x_3,x_4,x_5)$ constrained to satisfy the definition of an f-pentomino.

Concepts are specified in a logical language which supports composite terms (variables and constants; denoted here in bold). For example, given the concepts of an f-pentomino and a z-pentomino, we can define the composite concept of an f-pentomino-west-of-a-z-pentomino. This is specified in a higher order relation: $\texttt{f-pent-west-z-pent}(\mathbf{z}) \equiv \texttt{west}(\mathbf{x}, \mathbf{y}) \wedge \texttt{f-pent}(\textbf{y}) \wedge \texttt{z-pent}(\textbf{x})$. $\texttt{west}$ is taken to mean that all points $y_i$ are west of all points $x_i$. We define it as $\texttt{west}(\mathbf{x}, \mathbf{y}) \equiv [\bigwedge_{i,j} \texttt{west\_point}(x_i, y_j)] \wedge \textbf{x} == (x_i) \wedge \textbf{y} == (y_j)$ using the primitive concept $\texttt{west\_point}(x,y)$ which says that point $y$ is one grid point west of point $x$. Fig.~\ref{fig:f_pent}a illustrates the decomposition.


%
%


\begin{table}[!ht]
	\centering
	\resizebox{\columnwidth}{!}{%
	\begin{tabular}{c|c}
		\toprule
		\textbf{Concept}    &	\textbf{Definition}	\\
		\midrule
		red & $\texttt{red}(x) \equiv x.color = \texttt{RED}$ \\
		\midrule
		point & $\texttt{point}(x) \equiv$ $(\texttt{red}(x) \vee \texttt{blue}(x) \vee \cdots \vee \texttt{yellow}(x)) \wedge$ \\
		& $(x.x\_loc == 0 \vee \cdots \vee x.x\_loc == \texttt{GRID\_SIZE}-1) \wedge$ \\
		& $(x.y\_loc == 0 \vee \cdots \vee x.y\_loc == \texttt{GRID\_SIZE}-1)$ \\
		\midrule
		2x2 square & $\texttt{2x2\_square}(\mathbf{x}) \equiv \mathbf{x} == (x_1,x_2,x_3,x_4) \wedge \texttt{east}_1(x_1,x_2) \wedge $ \\ of points & $\texttt{south}_1(x_2,x_3) \wedge \texttt{west}_1(x_3,x_4)$ $\wedge \texttt{north}_1(x_3,x_1) \wedge$ \\ &
		$\texttt{point}(x_1) \wedge \texttt{point}(x_2)$ $\wedge \texttt{point}(x_3) \wedge \texttt{point}(x_4)$ \\
		\midrule
		south 1 point and same color & $\texttt{south}_1\texttt{\_same\_color}(x_1,x_2) \equiv \texttt{south}_1(x_1,x_2) \wedge x_1.color == x_2.color$\\
		\midrule
		f-pentomino &
		$\texttt{f-pent}(\mathbf{x}) \equiv$
		$\texttt{east}_1\_\texttt{same}\_\texttt{color}(x_1,x_2) \wedge \texttt{south}_1\_\texttt{same}\_\texttt{color}(x_4,x_1) \wedge$ \\ &  $\texttt{west}_1\_\texttt{same}\_\texttt{color}(x_4,x_3) \wedge \texttt{south}_1\_\texttt{same}\_\texttt{color}(x_4,x_5) \wedge \mathbf{x} == (x_1,x_2,x_3,x_4,x_5)$\\ 
		& $\wedge \texttt{point}(x_1) \wedge \texttt{point}(x_2) \wedge \texttt{point}(x_3) \wedge \texttt{point}(x_4) \wedge \texttt{point}(x_5)$ \\
		\bottomrule
	\end{tabular}
	\smallskip
	}
	\caption{Examples of concepts used in our study.}
	\label{tab:some-concepts}
\end{table}

\subsection{Generation: Concepts to Images}
\label{ssec:generation}
For each concept, we can generate a hierarchical scene graph~\cite{johnson2015image} whose lowest level corresponds directly to the image we want to generate. The details of the generation process can be found in Appendix 1. Here, we give a sketch of the procedure for the example shown in Figure \ref{fig:f_pent}. We don't discuss disjunction in this example but the algorithm can handle it by converting the concept definition to Disjunctive Normal Form (DNF) and applying this procedure to each conjunctive clause, until one is found which successfully generates.

For a unary concept like $\texttt{f\_pent\_west\_z\_pent}(\mathbf{z})$, we generate a single node for its object $\mathbf{z} = (\mathbf{x}, \mathbf{y})$ and recursively generate the constituent objects $\mathbf{x}$ and $\mathbf{y}$ according to the definition. This requires first building the scene graph for that definition, with nodes $\mathbf{x}$ and $\mathbf{y}$ and an edge between them labeled with the concept they satisfy, $\texttt{west}$. This graph is then traversed in breadth-first order starting from the first argument of the relation (here $\mathbf{x}$). First, the $\mathbf{x}$ node is recursively generated, followed by the $\texttt{west}$ relation, which recursively generates $\mathbf{y}$ so that it is west of $\mathbf{x}$. The $\texttt{f\_pent}$ is itself a composite on the points $y_1, y_2, y_3, y_4, y_5$, which constrains them using primitive concepts such as $\texttt{south\_1\_same\_color}$, which in turn is defined using the relational concepts ($==, \neq$, etc) on points.  The scene graph corresponding to the primitive concept level is an abstract representation of the grid and its points which can be directly rendered as an image for image classification experiments.

If the concept specification is inconsistent, the generator will fail and report an error. If the concept is disjunctive, another disjunct is chosen and the process is repeated until a successful graph can be generated. This generation procedure is therefore sensitive to disjunctions and may fail to terminate if there are too many. We timeout if that is the case, but all concepts in this paper generate quickly.

A full list of concepts, including examples of more complex concepts not used in this paper, is included in Appendix 1.
\section{Experimental Setup}
\label{sec:experiments}


The task for all experiments is multi-class prediction on images generated from a small set of concepts (3-5). Our goal is to better understand how standard as well as explicitly relational models generalize compositionally, for images with clear relational and compositional structure. We focus in particular on two specific types of compositional generalization: \emph{productivity} and \emph{substitutivity} \cite{hupkes2020compositionality}. The experiments on productivity (composition length generalization) test how well these models are able to learn recursive concepts if trained on a small number of compositions. In the substitutivity experiments, we test whether a model can learn to generalize correctly to other objects of the same type as its argument(s).\footnote{We will release the dataset upon publication.}

We choose an image size of 32x32 similar to CIFAR10~\cite{krizhevsky2009learning} and five colors: blue, red, green, yellow and white.
We compare an MLP; a 2-layers CNN; ResNet18~\cite{he2016deep}; PrediNet~\cite{shanahan2019explicitly}, which uses multi-head attention to extract relational structure; and WReN~\cite{santoro2018measuring}, which uses relation network modules~\cite{santoro2017simple}. These baselines were chosen to provide a balance between well-known architectures and recent ones with relational inductive biases. All models have approximately the same number of parameters for fair comparison, and we performed hyper-parameter optimization on each model, starting with published values for Predinet, ResNet and WReN.

When performing multiple tests with the same trained model, such as in Sec~\ref{ssec:alt-comp} where we test on sequence length 3 and 5, we keep the same test data for concepts which do not change. In Sec~\ref{ssec:alt-comp}, the "2x2 colored square" concept does not depend on sequence length: we use the same samples for both tests. The statistics for such classes therefore do not change between these variations. We report F1 score rather than accuracy as it better reflects performance on false positives and negatives. All numerical results shown in this section were averaged over 10 random seeds. We refer the reader to the appendix for more information (number of training samples etc.).

\section{Evaluation}
\subsection{Productivity Experiments: Generalization to Longer Compositions}
\label{ssec:comp-seq-gen}

\subsubsection{Pure and Mixed Sequences of 2x2 Squares}
\label{sssec:comp-single-concept}
This experiment evaluates the ability of a model to learn a recursive concept: $\texttt{east}_1(x_1,x_2)$, which requires at least one point in $x_2$ to be 1 grid point east of one point in $x_1$. We define 2x2 red or blue squares (see Table~\ref{tab:some-concepts}) and use them to construct horizontal sequences, by composing the $\texttt{east}_1$ concept on its first argument. We consider three sequence concepts: \textit{all red}, \textit{all blue}, and \textit{mixed red and blue}. We train on sequences of length 1 (a single red or blue square) and 2, then test on lengths 3, 5, and 7. Aggregated results are shown in Fig.~\ref{fig:exp_1_results}; per-concept results are available in Appendix 2.

\begin{figure}[!ht]
\centering
\begin{subfigure}{.4\textwidth}
  \centering
    \includegraphics[width=\textwidth]{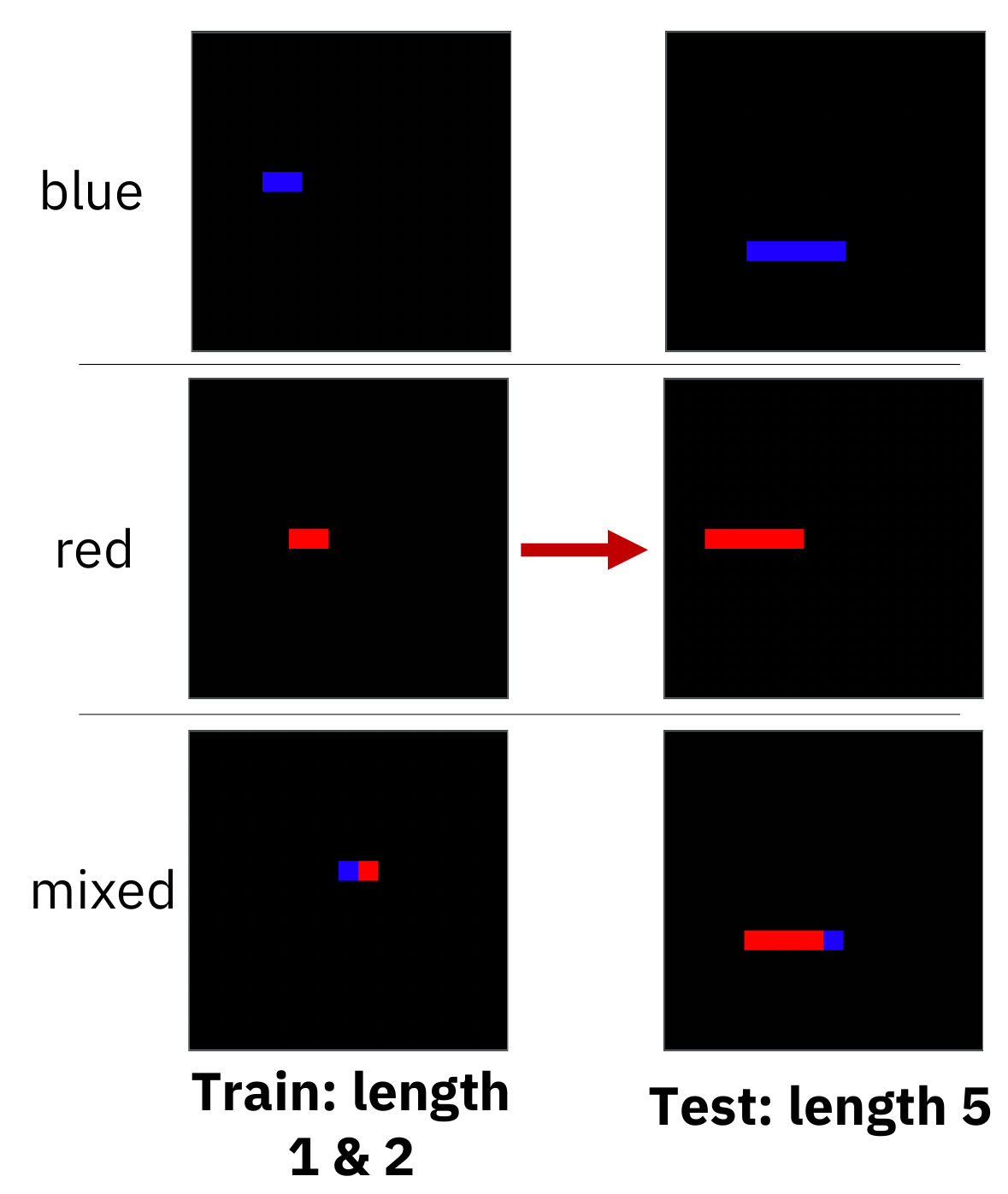}
  \caption{}
  \label{fig:exp_1_examples}
\end{subfigure}\hfill
\begin{subfigure}{.5\textwidth}
  \centering
    \includegraphics[width=\textwidth]{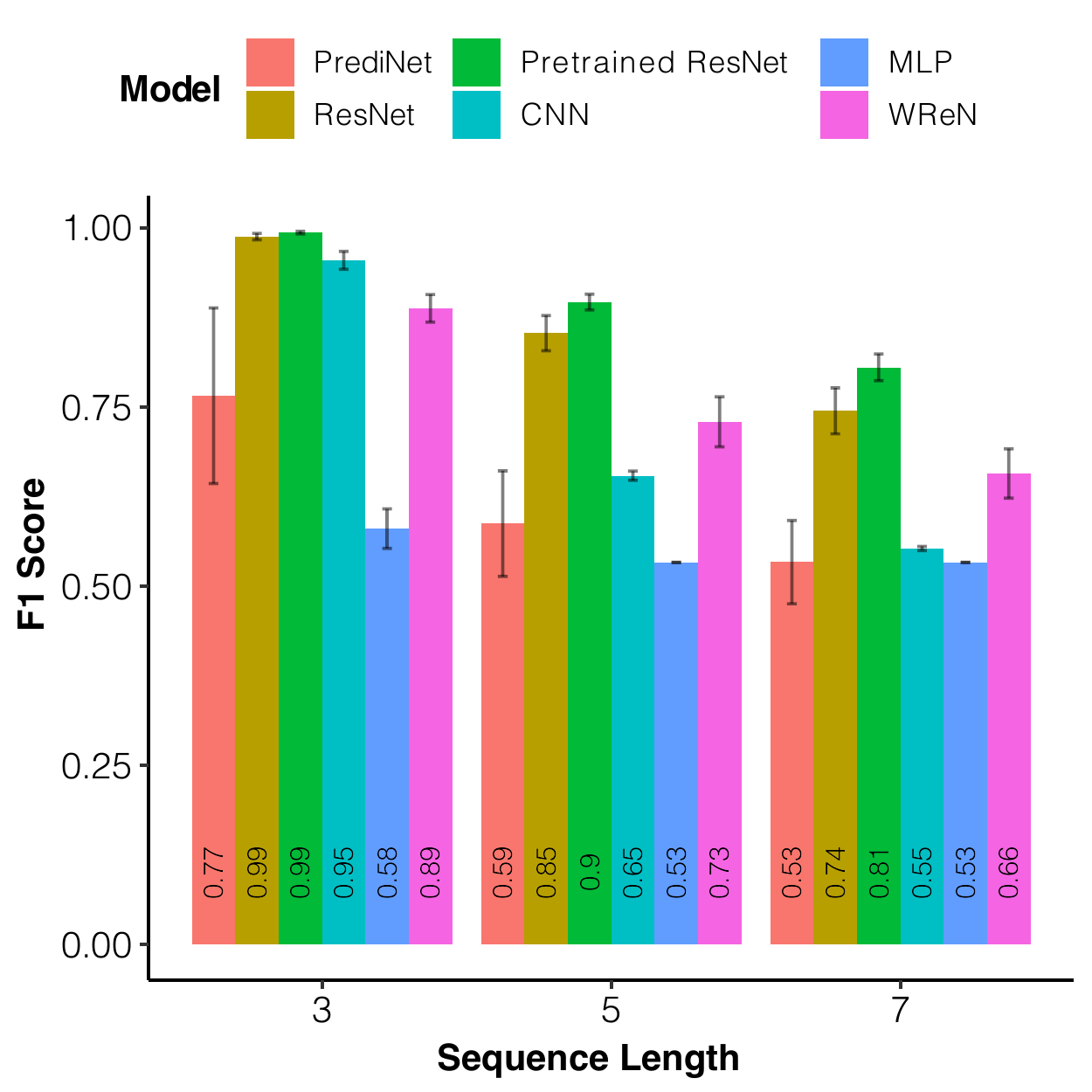}
    \caption{}
  \label{fig:exp_1_results}
\end{subfigure}
\caption{(a) Some train and test samples for experiment 1 (pure and mixed sequences, Sec.~\ref{sssec:comp-single-concept}). (b) F1 scores, aggregated over all concepts (sequence types). Error bars are 95\% CI over 10 seeds.}
\label{fig:exp_1}
\end{figure}

ResNet performs best overall but experiences a limited but noticeable degradation in generalization over longer sequences. Pretraining on ImageNet~\cite{deng2009imagenet} appears to be beneficial, particularly for longer sequences, presumably because pretraining helps the model learn more robust visual features which are helpful for the more complex test sequences. We use only the pretrained version of ResNet in subsequent experiments. The CNN, while architecturally simpler than ResNet, shows good performance but struggles on longer sequences. WReN, which uses a CNN to extract features, does better, particularly on mixed sequences. PrediNet performs similarly to the CNN, but suffers from high variance (we observed that PrediNet requires a significant amount of hyper-parameter tuning to reduce variance over different seeds). The MLP appears to have learned to average pixel values to classify the image.  As the mixed sequences contain both red and blue squares, the MLP cannot discriminate them and labels them as all red or all blue.

The sequences discussed here are spatial, constructed by composing the $\texttt{east}_1$ relation. To classify them correctly, models need to take the whole sequence into consideration, which is challenging as the length increases. While the WReN relational model performs reasonably well, PrediNet does poorly. The task here involves composing a single spatial relation and the images contain only a single object. In the next experiment, we try compositions involving both a spatial and color relation and multiple objects. 



\subsubsection{Box-World}
\label{ssec:alt-comp}

In this experiment, we recreate a simplified version of the Box-World environment of \cite{battaglia2018relational}, which is a grid-world domain with keys, locks and a gem. In the original reinforcement learning version of the game, the agent needs to find an initial (free) key and perform a series of "unlock" / "lock" steps, to find the gem. In our version, the task is to distinguish images with valid sequences from those with invalid, distractor, sequences, which can't be solved because the chain is broken somewhere with a key that doesn't open any locks. The problem is rendered as an image by representing keys, locks, and the gem as 2x2 squares (pixels); representing a paired lock and key by a common color; and a locked object (key or gem) as a square 1 pixel west of the lock. See Appendix 3 for an example.

This experiment tests a more complex sequence concept than experiment 1, as it requires learning to recurse on a conjunction of a spatial relation $\texttt{west}_1$ (for "locks") and non-spatial  $\texttt{same\_color}$ (for "unlocks"). 

The concepts to be distinguished here are $\texttt{solution}$ when a path leads to a gem, $\texttt{distractor}$ when a path does not lead to a gem, and $\texttt{2x2 square}$, the "pixel" or base element of the paths. We train the models with paths of length 1 and 2 and test on lengths 3 and 5. The results are shown in Fig.~\ref{fig:exp_3_results}. Full results are available in Appendix 3. 

All models, except PrediNet, retain the ability to identify the 2x2 square concept they were trained on. However, most models show poor generalization ability on learning longer $\texttt{solution}$ and $\texttt{distractor}$ sequences, even for length 3. WReN performs best (although not far from random), beating the pretrained ResNet, suggesting that it makes better use of the relational biases in its architecture. Nevertheless, it has difficulty on the $\texttt{distractor}$ concept. This is clearly visible in its confusion matrix (Fig.~\ref{fig:conf_matrix_wren}), where it can be seen that it mistakes most of the distractor paths for valid ones and mistakes fewer valid ones for distractor ones. 



\begin{figure}[!ht]
\centering
\begin{subfigure}{.5\textwidth}
  \centering
  \includegraphics[width=0.8\textwidth]{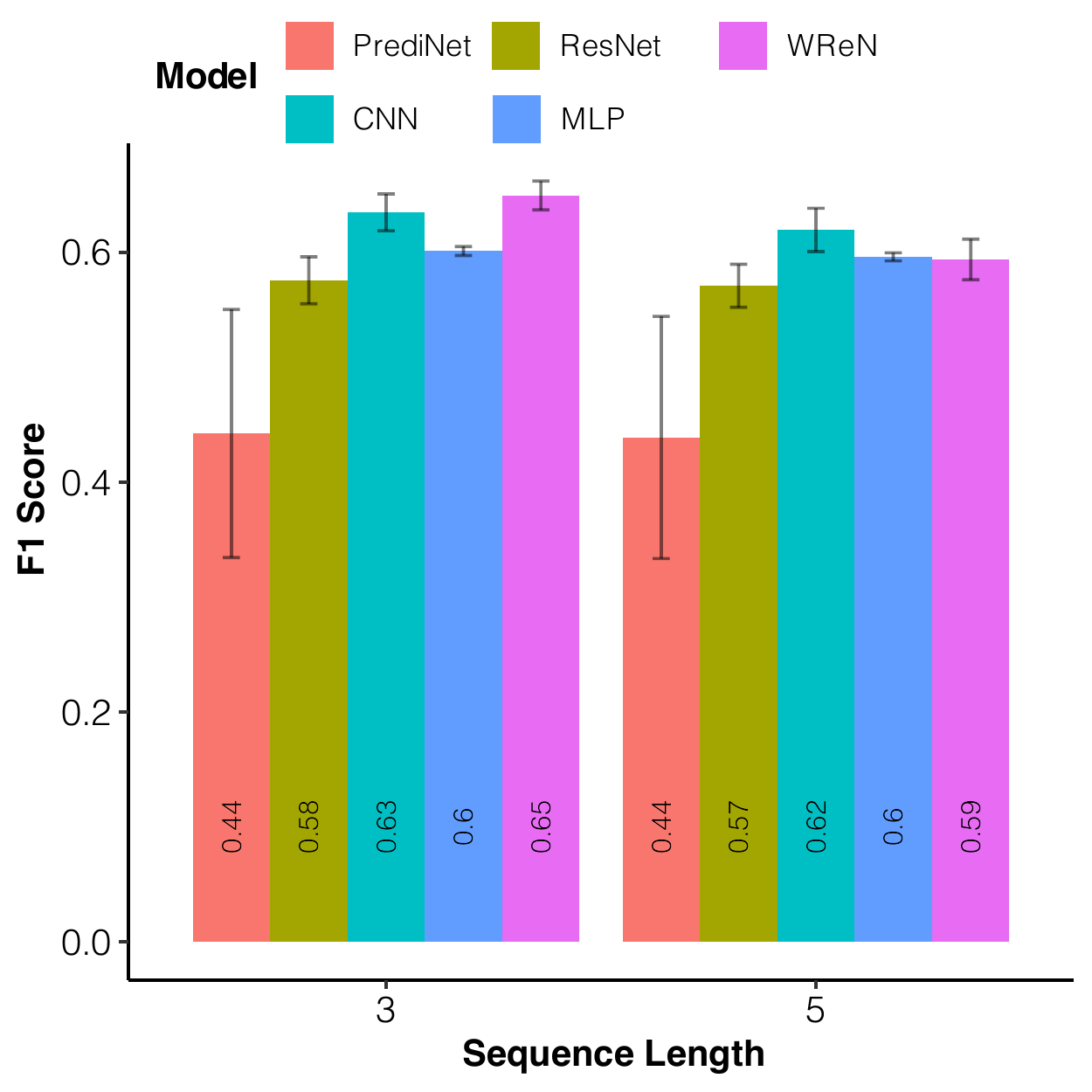}
    \caption{}
  \label{fig:exp_3_results}
\end{subfigure}\hfill
\begin{subfigure}{.5\textwidth}
  \centering
  \includegraphics[width=\textwidth]{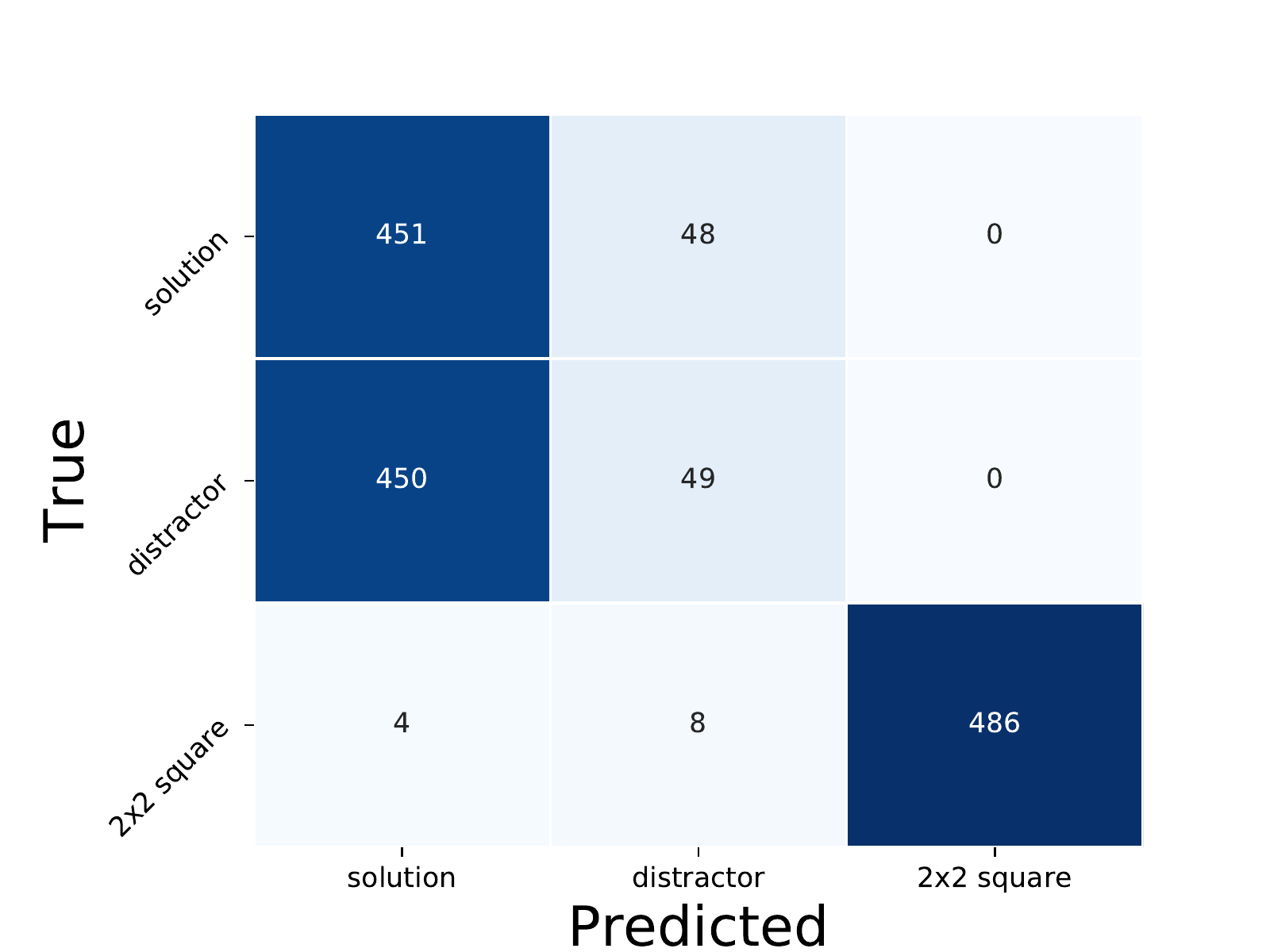}
  \caption{}
  \label{fig:conf_matrix_wren}
\end{subfigure}
\caption{Results for Experiment 2 (Box-World sequences). (a): Generalization (zero-shot transfer) over longer sequences of keys and locks. (b) Confusion Matrix of WReN on sequences of length 5 (averaged over 10 runs).}
\end{figure}

\subsection{Substitutivity Experiments: Scaling up 2x2 Patterned Squares}
\label{ssec:substitutivity}

In contrast to the previous two experiments where we test if models can learn to generalize to longer sequences of the \textit{same} object, the two experiments below test whether a concept generalizes to different objects (still satisfying the same concept) than seen in training. We consider variations on the number of arguments to substitute and the type of arguments to substitute.

\subsubsection{Generalizing from 2x2 to 4x4 patterned squares}
\label{sssec:visually-similar-substitute}


In this experiment, we create concepts which are 2x2 squares (see Table \ref{tab:some-concepts}) of four classes: \textit{all blue}, \textit{all red}, \textit{vertical alternating red/blue stripe}s, and a \textit{checkerboard pattern of red/blue}. The concept we want to test is that of a 2x2 square composed of 4 identical smaller squares -- one for each quadrant. For the training concepts, the "squares" being composed are 4 points (we consider these 1x1 squares here). We test whether a model can learn to generalize this relation by substituting 2x2 squares for the points creating a 2x2 square of 2x2 squares, which is a 4x4 square of points. The 2x2 squares to be substituted are the solid red and blue ones given in training. This substitution corresponds visually to "scaling up". We assign these scaled up 4x4 squares the same concept ids as the corresponding 2x2 versions. Figure \ref{fig:cf_exp_4_samples} shows some examples.  

We observe (table of F1 scores available in Appendix 4) that no model generalizes to the striped and checkered 4x4 squares. Among the models, ResNet, CNN and MLP are able to recognize the 4x4 all blue and all red squares after being trained on corresponding 2x2 ones. However, they aren't able to use the 2x2 squares compositionally to generalize to the 4x4 stripes or checkerboard. As an example, Fig.~\ref{fig:cf_exp_4_mlp} shows the MLP's confusion matrix, which suggests that it has learned a strategy that generalizes to some degree but not systematically.

\begin{figure}[!ht]
\centering
\begin{subfigure}{.5\textwidth}
  \centering
  \includegraphics[width=\textwidth]{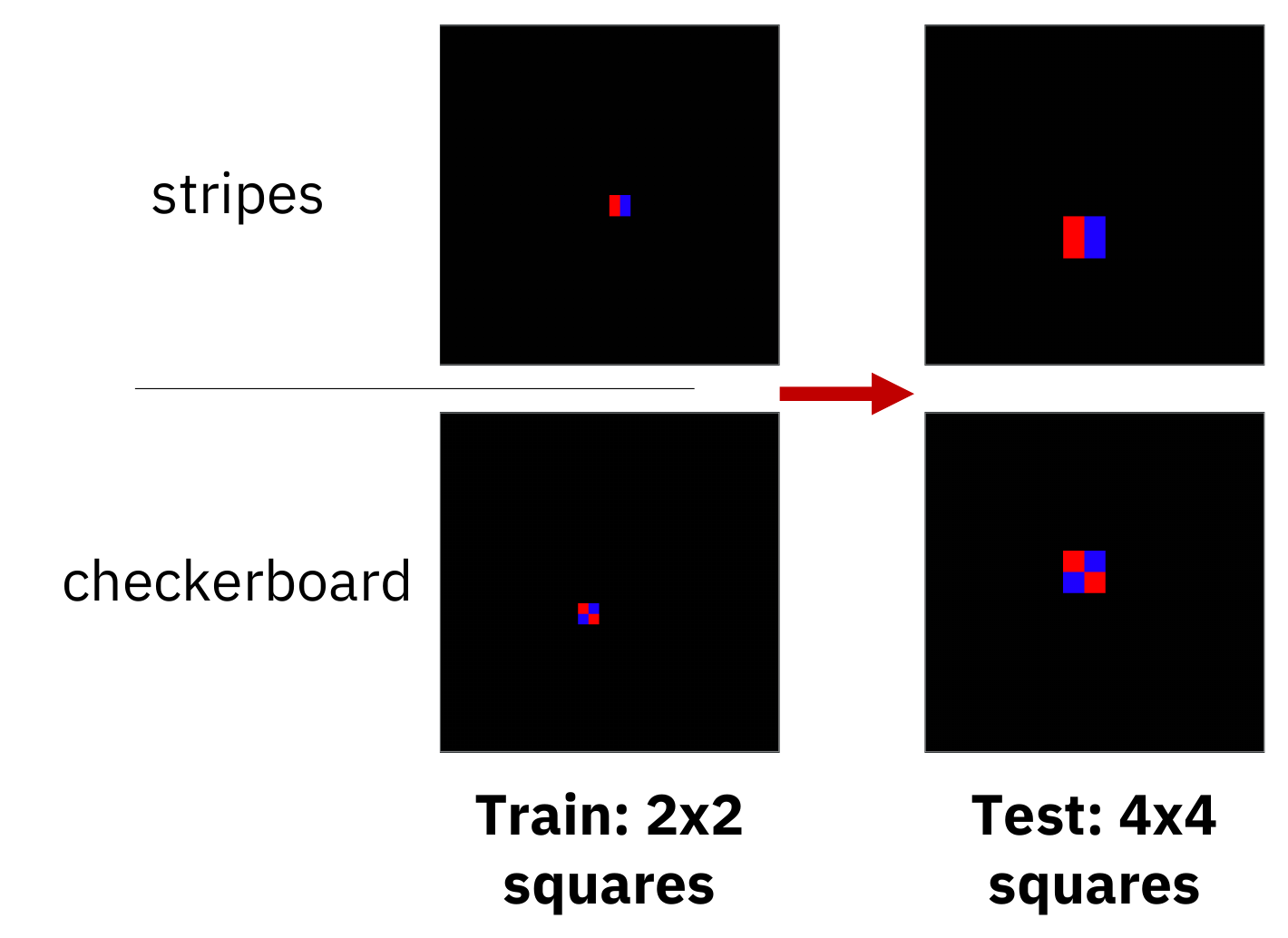}
    \caption{}
  \label{fig:cf_exp_4_samples}
\end{subfigure}\hfill
\begin{subfigure}{.5\textwidth}
  \centering
  \includegraphics[width=\textwidth]{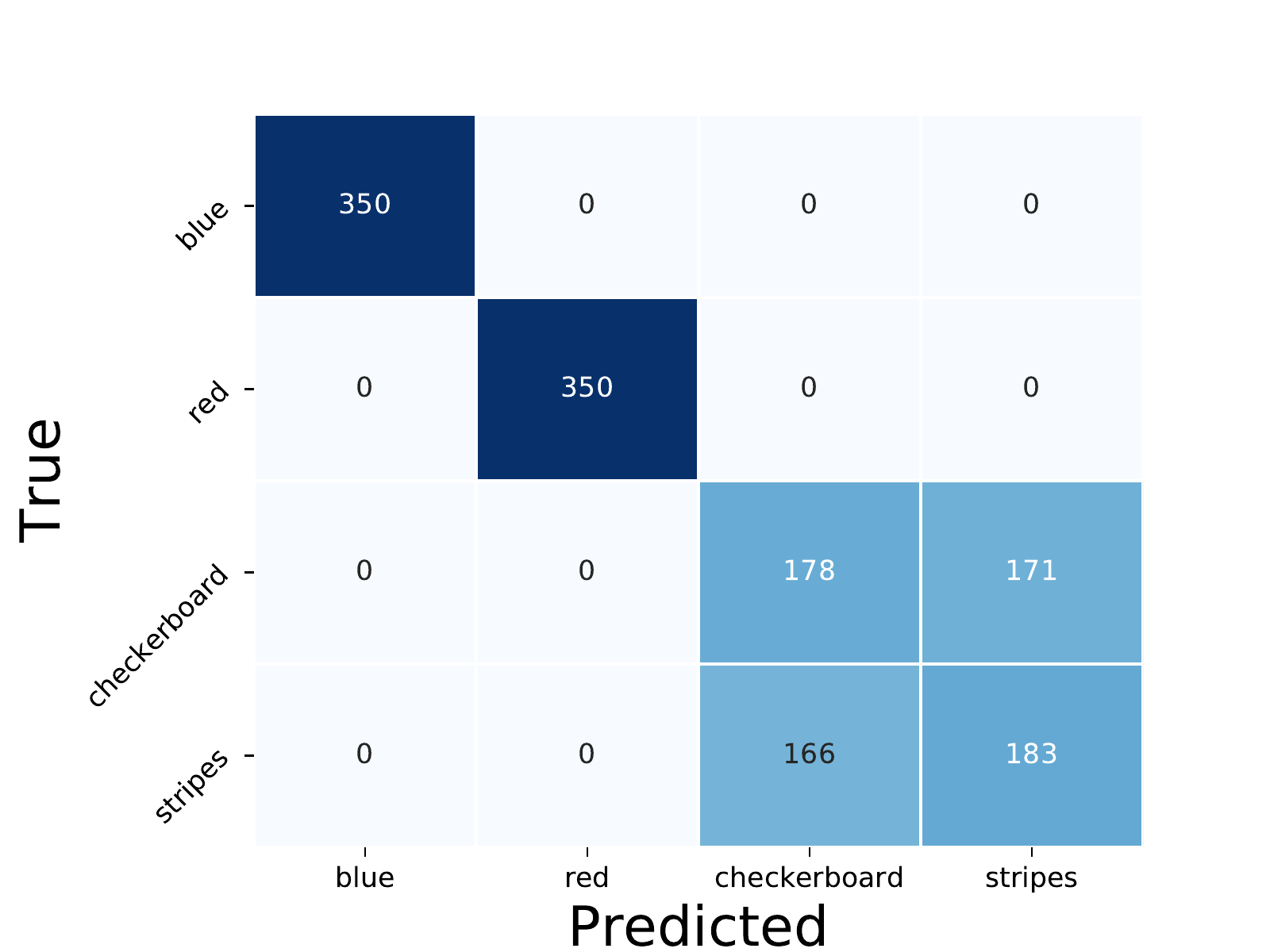}
  \caption{}
  \label{fig:cf_exp_4_mlp}
\end{subfigure}
\caption{(a) Train and test examples of the patterned squares concepts in experiment 3 (Sec.~\ref{sssec:visually-similar-substitute}). (b) Confusion matrix over the 4x4 squares concepts for the MLP model.}
\label{fig:cf_exp_4}
\end{figure}


\subsubsection{Generalizing sequences under element substitutions}
\label{sssec:visually-different-substitute}
\begin{figure}[!ht]
\centering
\begin{subfigure}{.6\textwidth}
  \centering
  \includegraphics[width=\textwidth]{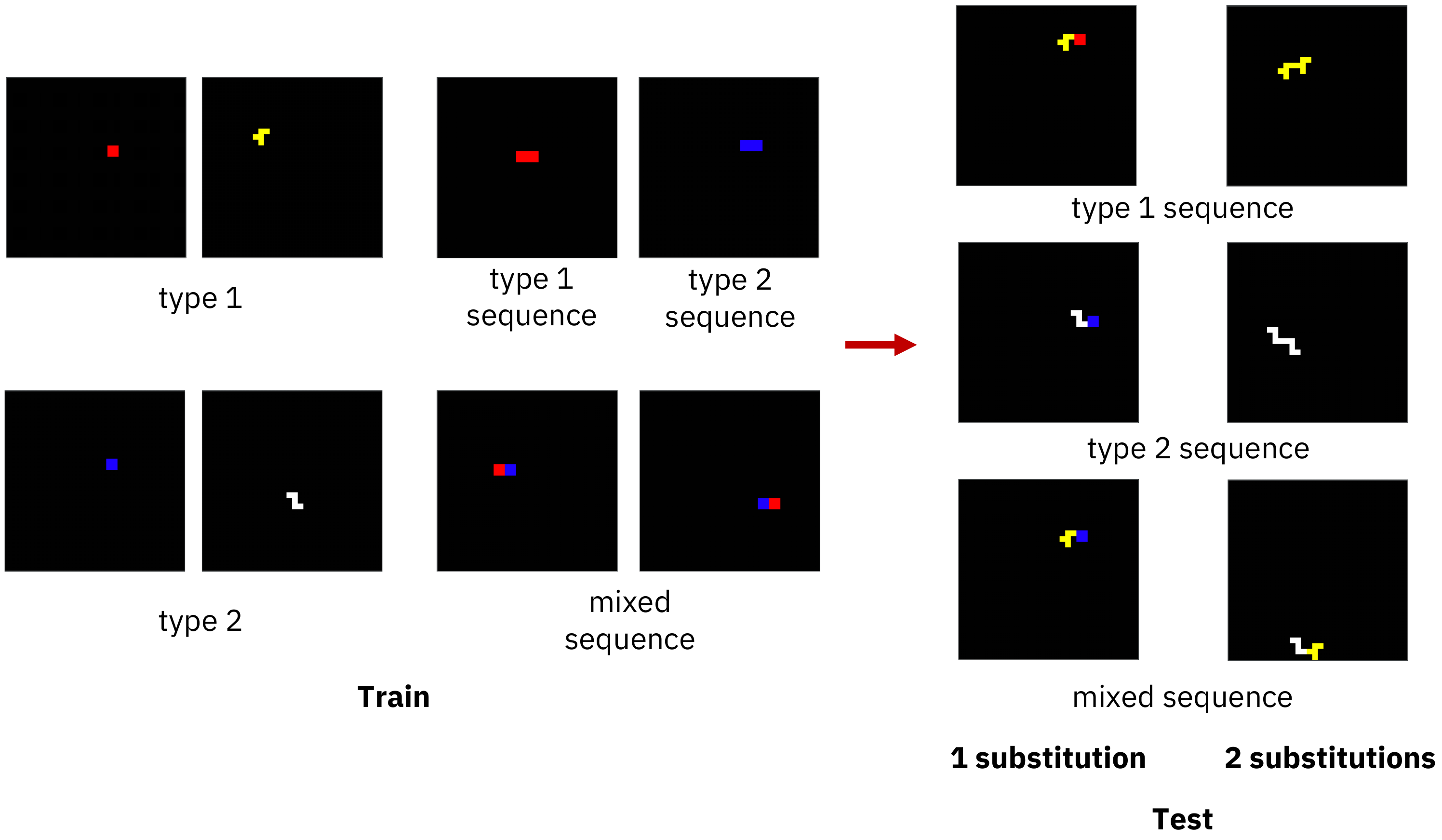}
  \caption{}
  \label{fig:cf_exp_5_samples}
\end{subfigure}\hfill
\begin{subfigure}{.4\textwidth}
  \centering
    \includegraphics[width=\textwidth]{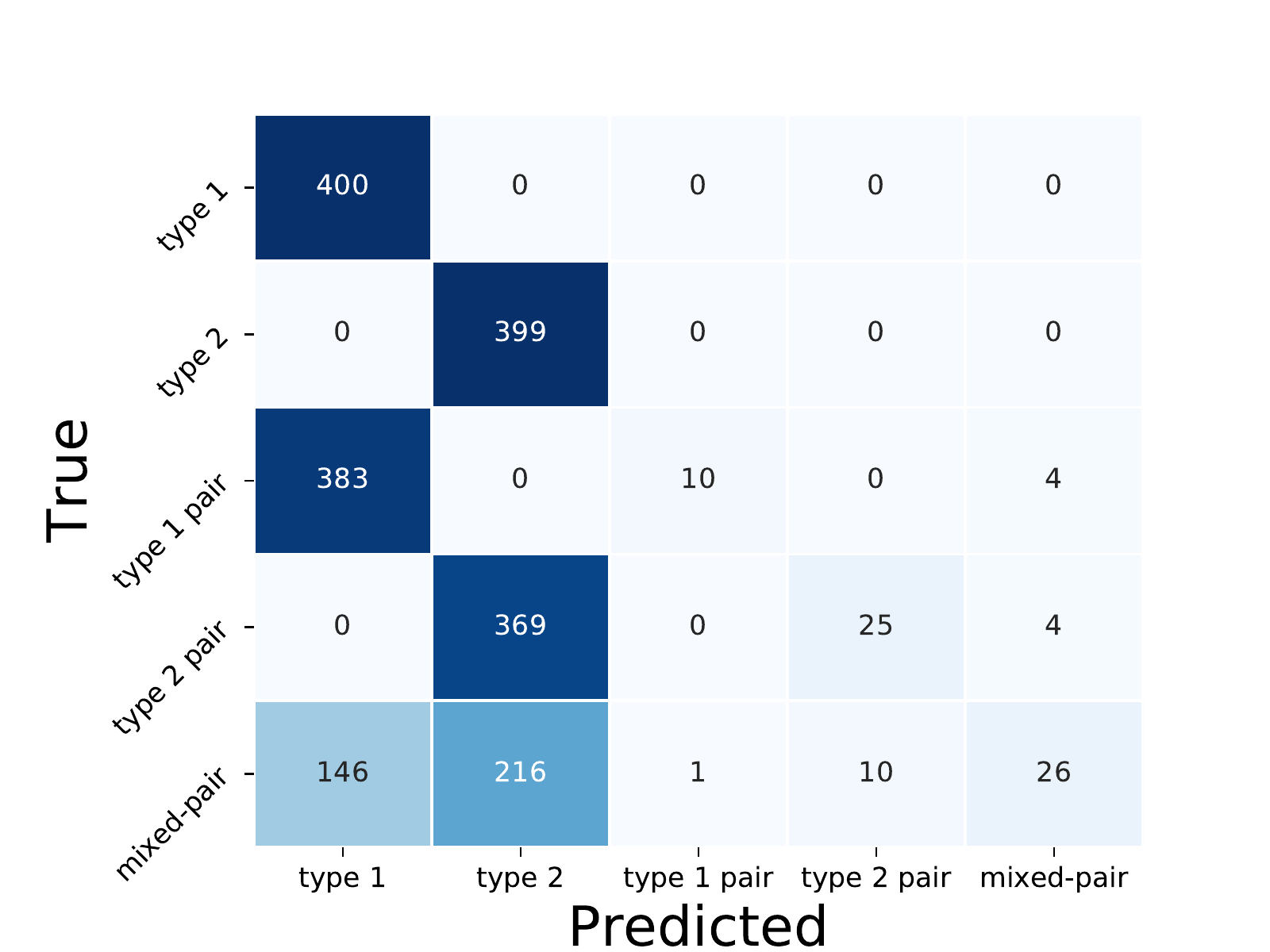}
    \caption{}
  \label{fig:cf_exp_5_resnet}

\end{subfigure}
\caption{Experiment 4 (sequence substitutions, Sec.~\ref{sssec:visually-different-substitute}): (a) Some train and test samples of the types and sequences. (b) Average confusion matrix over 1-substitute pairs for ResNet: it recognizes true \texttt{type 1} or \texttt{type 2} examples, but also labels as such the pairs.}
\label{fig:cf_exp_5}
\end{figure}

In this experiment, we test whether a model trained on a concept $\texttt{c}$ and sequences of instances $e \in \texttt{c}$, can generalize to sequences of elements $e' \in \texttt{c}$ for $e \neq e'$. Figure \ref{fig:cf_exp_5_samples} shows examples where the element class is concept $\texttt{type 1}$ which consists of a 2x2 red square and an f-pentomino; or $\texttt{type 2}$ which consists of a blue square and a z-pentomino. We evaluate the ability of the model to discriminate three types of sequences of these elements: pure $\texttt{type 1}$, pure $\texttt{type 2}$, and $\texttt{mixed 1+2}$. The model receives training on $\texttt{Type 1}$, $\texttt{Type 2}$ and sequences involving squares from these classes. It must generalize to classify sequences involving \textit{both} pentominoes and squares.  There are five classes in total including the element types and sequences. For the pentominoes, color is ignored and only type matters.

We test generalization of the sequences as follows. A $\texttt{type 1}$ sequence is created with both a red square and an f-pentomino; a $\texttt{type 2}$ sequence is created with both a blue square and a z-pentomino; and a mixed sequence with a blue square and f-pentomino or a red square and z-pentomino. This tests the ability of the model to perform one substitution in the sequence. We also evaluate 2 substitutions, by changing both elements of the sequence. A $\texttt{type 1}$ sequence is an f-pent pair; a $\texttt{type 2}$ sequence is a z-pent pair; and a $\texttt{mixed 1+2}$ sequence is an f-z pair.

To solve this problem robustly, the models need to learn to associate visually different objects as a common class and generalize a concept trained on some elements of this class to correctly classify new objects of the class not seen in training.

The models generally learn to recognize the \texttt{type 1} and \texttt{type 2} classes but with low precision, as they confuse the pairs for the single arguments. This is particularly visible with ResNet (Fig.~\ref{fig:cf_exp_5_resnet}). One issue could be that learning the \texttt{type 1} and \texttt{type 2} concepts concurrently with the higher-order concepts may be too difficult. We tested a curriculum variant, where we trained until convergence on \texttt{type 1} and \texttt{type 2}, then added the remaining 3 sequence concepts. Except for ResNet, which improved slightly, the models all degraded. We hypothesize that a curriculum is not helpful without a composition mechanism to properly make use of it. Interestingly, while the relational models PrediNet and WReN do not surpass other models on the \texttt{type 1} and \texttt{type 2} classes, they perform relatively better on the sequences. Absolute performance remains poor.

\subsection{Discussion}
All productivity experiments show degradation in performance of all models as composition length increases, indicating they have not learned a recursive generalization. For the Box-World experiment, no models do well (F1 scores are around 0.6 on sequence length 3 and performance degrades slightly for length 5). We note however that the pretrained ResNet performs well, only to be surpassed by WReN on the Box-World solution paths. For the relational models, WReN performs reasonably well but PrediNet does poorly. This indicates to us that their relational bias is perhaps not helpful without an additional compositional bias to employ the learned relations recursively. For the substitutivity experiments, while all models fail to properly generalize, WReN and PrediNet perform relatively better with the elements substitutions. These tasks are qualitatively harder and seem to require a more explicit compositional bias to generalize at all.


\section{Conclusion}
\label{sec:conclusion}

In this paper, we investigated the problem of compositional, relational generalization as a multi-class image recognition problem. We introduced a concept specification language, which allows a description of hierarchical concepts on a grid of colored points, and sketched the generation process that renders them into images. We found that having a declarative definition of the concepts facilitated more agile experimentation and concept sharing. Our experiments on compositional productivity and substitutivity provide evidence that, without specific biases for both relation representation as well as composition, neural networks do not generalize well in this setting and suffer from the same degradation on composition length as seen in experiments on text and multi-modal data.

\newpage
\bibliographystyle{alpha}
\bibliography{bibliography}

\newpage
\setcounter{section}{0} 

\section{List of Concepts and Details of the Generation Algorithm}
\label{app:concepts}

\subsection{List of concepts}
Table~\ref{tab:full_concepts} contains the definition of the concepts we used in our experiments. For vector objects, we use the notation $x::\mathbf{y}$ to mean the function which splits the vector into its first element ($x$) and the rest of the elements ($\mathbf{y}$). More generally, we can split off $k$ elements from the front of the sequence by writing $(x_1,x_2,\cdots,x_k)::\mathbf{y}$. We write singleton vectors $\mathbf{x}$ as $\mathbf{x} == (x)$. Quantification over vectors is interpreted to mean over the set of points of the vector, forgetting the structure. We allow two reduction operations on integer-valued properties of a vector: $\argmin{}$ and $\argmax{}$. For example, $\argmin_{x \in \mathbf{x}}(x.x\_loc)$ produces the the element $x$ in $\mathbf{x}$ for which $x.x\_loc$ is minimal. We have defined $\texttt{east}_1$ to use these functions rather than conjunction (as in the main paper) for increased efficiency in generation.

\begin{table}[!ht]
	\centering
	\resizebox{\columnwidth}{!}{%
	\begin{tabular}{c|c}
		\toprule
		\textbf{Concept}    &	\textbf{Definition}	\\
		\midrule
		point & $\texttt{point}(x) \equiv (\texttt{red}(x) \vee \texttt{blue}(x) \vee \cdots \vee \texttt{yellow}(x)) \wedge$ \\
		& $(x.x\_loc == 0 \vee \cdots \vee x.x\_loc == \texttt{GRID\_SIZE}-1) \wedge$ \\
		& $(x.y\_loc == 0 \vee \cdots \vee x.y\_loc == \texttt{GRID\_SIZE}-1)$ \\
		\midrule
		red (point) & $\texttt{red\_point}(x) \equiv \texttt{point}(x) \wedge x.color == \texttt{RED}$ \\
		\midrule
		red (object) & $\texttt{red}(\mathbf{x}) \equiv \forall{x} \in \mathbf{x} \; \texttt{red\_point}(x)$ \\
		\midrule
		object with red or blue points & $\texttt{red\_or\_blue}(\mathbf{x}) \equiv \forall{x} \in \mathbf{x} \; [\texttt{red}(x) \vee \texttt{blue}(x)]$\\
		\midrule
		same color (point) & $\texttt{same\_color\_point}(x_1,x_2) \equiv x_1.color == x_2.color \wedge \texttt{point}(x_1) \wedge \texttt{point}(x_2)$\\
		\midrule
		east 1 grid point (point) & $\texttt{east\_point}_1(x_1,x_2) \equiv x_2.x\_loc == x_1.x\_loc+1 \wedge$ \\ & $x_2.y\_loc == x_1.y\_loc \wedge \texttt{point}(x_1) \wedge \texttt{point}(x_2)$ \\
		\midrule
		east (point) & $\texttt{east\_point}(x_1,x_2) \equiv x_2.x\_loc > x_1.x\_loc \wedge x_2.y\_loc == x_1.y\_loc$\\
		\midrule
		east (object) & $\texttt{east}(\mathbf{x}_1,\mathbf{x}_2) \equiv x_1 == \argmax_{x \in \mathbf{x_1}}\{{x.x\_loc}\} \wedge x_2 == \argmin_{y \in \mathbf{x_2}}\{{y.x\_loc}\} \wedge \texttt{east\_point}(x_1,x_2)$\\
		\midrule
		east 1 grid point (object) &
		$\texttt{east}_1(\mathbf{x}_1,\mathbf{x}_2) \equiv \texttt{east}(\mathbf{x}_1,\mathbf{x}_2)  \wedge \exists{x} \in \mathbf{x}_1, y \in \mathbf{x}_2\; \texttt{east\_point}_1(x,y)$\\
		\midrule
		east 1 sequence & $\texttt{east\_seq}(\mathbf{x}) \equiv [\mathbf{x} == (x)] \vee [\mathbf{x} == x::\texttt{xs} \wedge \texttt{xs} == y::\texttt{ys} \wedge \texttt{east\_point}_1(x,y) \wedge \texttt{east\_seq}(\texttt{xs})]$\\
		\midrule
		red east sequence (Exp 1) & $\texttt{red\_east\_seq}(\mathbf{x}) \equiv \texttt{east\_seq}(\mathbf{x}) \wedge \texttt{red}(\mathbf{x})$\\
		\midrule
		red or blue east sequence (Exp 1) & $\texttt{red\_or\_blue\_east\_seq}(\mathbf{x}) \equiv \texttt{east\_seq}(\mathbf{x}) \wedge \texttt{red\_or\_blue}(\mathbf{x})$\\
		\midrule
		north 1 grid point (point) & $\texttt{north\_point}_1(x_1,x_2) \equiv x_2.y\_loc == x_1.y\_loc-1 \wedge x_2.x\_loc == x_1.x\_loc$ \\ & $\wedge \texttt{point}(x_1) \wedge \texttt{point}(x_2)$ \\
		\midrule
		north 1 grid point (object) &
		$\texttt{north}_1(\mathbf{x},\mathbf{y}) \equiv \forall{x} \in \mathbf{x}, y \in \mathbf{y} \; \texttt{north\_point}_1(x,y)$\\
		\midrule
		2x2 square of points (Exp 1,2,3,4) & $\texttt{2x2\_square\_point}(\mathbf{x}) \equiv \texttt{east}_1(x_1,x_2) \wedge \texttt{south}_1(x_2,x_3) \wedge \texttt{west}_1(x_3,x_4) \wedge$ $\texttt{north}_1(x_4,x_1) \wedge$ \\ & $\mathbf{x} == (x_1,x_2,x_3,x_4) \wedge \texttt{point}(x_1) \wedge \texttt{point}(x_2) \wedge \texttt{point}(x_3) \wedge \texttt{point}(x_4)$ \\
		\midrule
		2x2 checkerboard (points) (Exp 3) & $\texttt{2x2\_checkerboard}(\mathbf{x}) \equiv \texttt{2x2\_square\_point}(\mathbf{x}) \wedge \mathbf{x} == (x_1,x_2,x_3,x_4) \wedge \texttt{red}(x_1) \wedge$\\ & $\texttt{blue}(x_2) \wedge \texttt{red}(x_3) \wedge \texttt{blue}(x_4)$ \\
		\midrule
		2x2 vertical stripes (points) (Exp 3) & $\texttt{2x2\_vert\_stripe}(\mathbf{x}) \equiv \texttt{2x2\_square\_point}(\mathbf{x}) \wedge \mathbf{x} = (x_1,x_2,x_3,x_4)$\\ & $\wedge \texttt{red}(x_1) \wedge \texttt{blue}(x_2) \wedge \texttt{blue}(x_3) \wedge \texttt{red}(x_4)$ \\
		\midrule
		square shape & $\texttt{square}(\mathbf{x}) \equiv \mathbf{x} == (x_1,x_2,x_3,x_4) \wedge \texttt{point}(x_1) \wedge\texttt{point}(x_2) \wedge \texttt{point}(x_3) \wedge \texttt{point}(x_4) \wedge$\\
		& $\texttt{east}(x_1,x_2) \wedge \texttt{south}(x_2,x_3) \wedge \texttt{west}(x_3,x_4) \wedge \texttt{north}(x_4,x_1)$\\
		\midrule
		2x2 square of squares (Exp 3) &
		$\texttt{2x2\_square\_of\_squares}(\mathbf{x}) \equiv \mathbf{x} = (\mathbf{x_1},\mathbf{x_2},\mathbf{x_3},\mathbf{x_4}) \wedge$ \\ & $\texttt{square}(\mathbf{x_1}) \wedge \texttt{square}(\mathbf{x_2}) \wedge \texttt{square}(\mathbf{x_3}) \wedge \texttt{square}(\mathbf{x_4}) \wedge$ \\
		 & $\mathbf{x_1} == (ul_1,ur_1,lr_1,ll_1) \wedge \mathbf{x_2} == (ul_2,ur_2,lr_2,ll_2) \wedge$\\
		 & $\mathbf{x_3} == (ul_3,ur_3,lr_3,ll_3) \wedge \mathbf{x_4} == (ul_4,ur_4,lr_4,ll_4) \wedge$ \\ & $\texttt{east}_1(ur_1,ul_2) \wedge \texttt{east}_1(lr_1,ll_2) \wedge$  $\texttt{south}_1(ll_2,ul_3) \wedge$ \\ & $\texttt{south}_1(lr_2,ur_3) \wedge \texttt{west}_1(ul_3,ur_4) \wedge \texttt{west}_1(ll_3,lr_4)$ \\
		\midrule
		2x2 square of squares & $\texttt{2x2\_square\_of\_squares\_checkerboard}(\mathbf{x}) \equiv \texttt{2x2\_square\_of\_squares}(\mathbf{x}) \wedge$ \\
		checkerboard (Exp 3) & $\mathbf{x} == (\mathbf{x}_1,\mathbf{x}_2,\mathbf{x}_3,\mathbf{x}_4) \wedge \texttt{red}(\mathbf{x}_1) \wedge \texttt{blue}(\mathbf{x}_2) \wedge \texttt{red}(\mathbf{x}_3) \wedge \texttt{blue}(\mathbf{x}_4)$ \\
		\midrule
		adjacency & $\texttt{adj}(x_1,x_2) \equiv \texttt{north}_1(x_1,x_2) \vee \texttt{east}_1(x_1,x_2) \vee \texttt{south}_1(x_1,x_2) \vee \texttt{west}_1(x_1,x_2)$ \\
		\midrule
		pentomino &
		$\texttt{pentomino}(\mathbf{x}) \equiv \texttt{adj}\_\texttt{same}\_\texttt{color}(x_1,x_2) \wedge \texttt{adj}\_\texttt{same}\_\texttt{color}(x_2,x_3) \wedge$ \\
		& $\texttt{adj}\_\texttt{same}\_\texttt{color}(x_3,x_4) \wedge \texttt{adj}\_\texttt{same}\_\texttt{color}(x_4,x_5) \wedge \mathbf{x} == (x_1,x_2,x_3,x_4,x_5)$ \\
		\midrule
		"f"-pentomino (Exp 4) &
		$\texttt{f-pent}(\mathbf{x}) \equiv$
		$\texttt{east}_1\_\texttt{same}\_\texttt{color}(x_1,x_2) \wedge \texttt{south}_1\_\texttt{same}\_\texttt{color}(x_4,x_1) \wedge$ \\ &  $\texttt{west}_1\_\texttt{same}\_\texttt{color}(x_4,x_3) \wedge \texttt{south}_1\_\texttt{same}\_\texttt{color}(x_4,x_5) \wedge \mathbf{x} == (x_1,x_2,x_3,x_4,x_5)$\\ 
		& $\wedge \texttt{point}(x_1) \wedge \texttt{point}(x_2) \wedge \texttt{point}(x_3) \wedge \texttt{point}(x_4) \wedge \texttt{point}(x_5)$ \\
		\midrule
		``z" pentomino (Exp 4) &
		$\texttt{z-pent}(\mathbf{x}) \equiv$
		$\texttt{east}_1\_\texttt{same}\_\texttt{color}(x_1,x_2) \wedge \texttt{south}_1\_\texttt{same}\_\texttt{color}(x_2,x_3) \wedge$ \\ & $\texttt{south}_1\_\texttt{same}\_\texttt{color}(x_3,x_4) \wedge \texttt{east}_1\_\texttt{same}\_\texttt{color}(x_4,x_5) \wedge \mathbf{x} == (x_1,x_2,x_3,x_4,x_5)$ \\ & $\wedge \texttt{point}(x_1) \wedge \texttt{point}(x_2) \wedge \texttt{point}(x_3) \wedge \texttt{point}(x_4) \wedge \texttt{point}(x_5)$ \\
		\midrule
		key & $\texttt{key}(\mathbf{x}) \equiv \texttt{2x2\_square\_point}(\mathbf{x})$; \\ lock & $\texttt{lock}(\mathbf{x}) \equiv \texttt{2x2\_square\_point}(\mathbf{x})$; \\ gem &  $\texttt{gem}(\mathbf{x}) \equiv \texttt{2x2\_square\_point}(\mathbf{x})$\\
		\midrule
		key or gem & $\texttt{key\_or\_gem}(\mathbf{x}) \equiv \texttt{key}(\mathbf{x}) \vee \texttt{gem}(\mathbf{x})$\\
		\midrule
		locked object & $\texttt{locks}(\mathbf{l},\mathbf{o}) \equiv \texttt{key\_or\_gem}(\mathbf{o}) \wedge \texttt{lock}(\mathbf{l}) \wedge \texttt{east}_1(\mathbf{o},\mathbf{l})$ \\
		\midrule
        key unlocks lock & $\texttt{unlocks\_lock}(\mathbf{k},\mathbf{l}) \equiv \texttt{key}(\mathbf{k}) \wedge \texttt{lock}(\mathbf{k}) \wedge \texttt{same\_color}(\mathbf{k},\mathbf{l})$\\
        \midrule
        solution (Exp 2) & $\texttt{solution}(\mathbf{s}) \equiv [\mathbf{s} == (\mathbf{g}) \wedge \texttt{gem}(\mathbf{g})] \vee$\\
        & $\mathbf{s} == \mathbf{k}::\mathbf{s}_1 \wedge \mathbf{s}_1 == \mathbf{l}::\mathbf{s}_2 \wedge \texttt{unlocks\_lock}(\mathbf{k},\mathbf{l}) \wedge \texttt{solution}(\mathbf{s}_2)$\\
        \midrule
        distractor position 2 (Exp 2) & $\texttt{distractor}(\mathbf{s}) \equiv \mathbf{s} == (\mathbf{k}_1,\mathbf{l}_1)::\mathbf{s}_1 \wedge$\\
        & $\texttt{unlocks\_lock}(\mathbf{k}_1,\mathbf{l}_1) \wedge \mathbf{s}_1 == (\mathbf{k}_2,\mathbf{l}_2)::\mathbf{s}_2 \wedge$\\
        & $\texttt{locks}(\mathbf{l}_1,\mathbf{k}_2) \wedge \texttt{not\_unlocks\_lock}(\mathbf{k}_2,\mathbf{l}_2) \wedge \texttt{solution}
        (\mathbf{s}_2)$\\
        \midrule
        composite x pentomino (i,u,f,z,x) Appx Fig 1.(c) & $\texttt{composite\_x\_pent}(\mathbf{x}) \equiv \texttt{x-pent}(\mathbf{x}) \wedge \mathbf{x} == (\mathbf{x}_1,\mathbf{x}_2,\mathbf{x}_3,\mathbf{x}_4,\mathbf{x}_5) \wedge$\\
        & $\texttt{i-pent}(\mathbf{x}_1) \wedge \texttt{u-pent}(\mathbf{x}_2) \wedge \texttt{f-pent}(\mathbf{x}_3) \wedge \texttt{z-pent}(\mathbf{x}_4) \wedge \texttt{x-pent}(\mathbf{x}_5) \wedge$ \\ & $\texttt{south}_1(\mathbf{x}_1,\mathbf{x}_3) \wedge \texttt{west}_1(\mathbf{x}_3,\mathbf{x}_2)\wedge \texttt{east}_1(\mathbf{x}_3,\mathbf{x}_4) \wedge \texttt{south}_1(\mathbf{x}_3,\mathbf{x}_5)$\\
		\bottomrule
	\end{tabular}
	}
	\smallskip
	\caption{The definition of the concepts used in our experiments. The left column contains a short textual description and in which experiment(s) was the concept used.}
	\label{tab:full_concepts}
\end{table}

\clearpage
\subsection{Details of the Generation Algorithm}

For clarity, we have separated the description of the algorithms in several parts, from high-level structure (Fig.~\ref{alg:gen-alg}) to specific functions (Fig.~\ref{alg:specific-functions}).

\begin{figure}[!ht]
    \begin{algorithm}[H]
        \SetKwProg{try}{try}{}{}
        \SetKwProg{catch}{catch}{}{end}
        \DontPrintSemicolon
        \SetNoFillComment
        \KwIn{\textup{c} : \textup{Concept}}
        \tcc{A Concept has: (1) a name; (2) an argument; (3) a definition.\;
        we assume a single arg (unary) because
        if a concept has more we can always \;
        wrap them up as a vector at a higher level. Example: \;
        c(x) := x=(x1,x2) and red(x\_1) and blue(x\_2)
        here the name is "c" \;
        the argument is "x" and the definition is "red(x\_1) and blue(x\_2).\;
        The constituent elements of x are x1 and 2.} \;
        
        \tcc{create a map of variables to their bindings,\;
            which are vectors of their constituent variables or point objects}
        $\var{bindings} = \var{Map}()$ \;
        \texttt{\\}
        \tcc{convert concept to disjunctive normal form (DNF) with an OR of conjunctive concepts}
        $\texttt{conj\_concepts} = \texttt{convert\_to\_dnf}(\texttt{c})$ \;
        \texttt{\\}
        \tcc{generate each conjunctive concept. if we fail; try another}
        \ForAll{\textup{conj\_concept} $\in$ \textup{conj\_concepts}}{
        \try{} {
          $\texttt{generate\_conjunctive}(\texttt{conj\_concepts}, \texttt{bindings})$
        }
        \catch {} {
          $\texttt{continue}$
        }
    }
    \caption{\texttt{generate\_concept}}
    \end{algorithm}
    \caption{Generation Algorithm: Iterate over the different conjunctive clauses of a concept until one can be successfully generated.}
    \label{alg:gen-alg}
\end{figure}

\begin{figure}[!ht]
    \begin{algorithm}[H]
        \KwIn{c : Concept; bindings : Map}
        \DontPrintSemicolon
        \SetNoFillComment
        \tcc{create a graph from the conjunctive concept definition: \;
            variables become vertices; unary relations become types associated with the vertex;\;
            binary relations become edges between vertices.\;
            We assume connectivity here but it's not a hard requirement.\;
            For simplicity, we assume one unary relation on any node and at most one edge between nodes.\;
            If there are, for example, 2 unary relations on a node, say c(x), d(x), then these can be \;
            grouped as e(x) := c(x) $\wedge$ d(x). Similarly for binary relations. So this is w.l.o.g.\;
            }
        $\var{g} = \func{create\_graph}(\texttt{c})$ \;
        \texttt{\\}
        \tcc{pick a variable to act as the root of the search,\;
            chosen randomly but bound variables are prioritized.}
        $\var{root\_variable} = \func{pick\_root}(\var{g})$ \;
         \texttt{\\}
        \tcc{get the root variable concept and the vector of variables used \;
        in the root variable concept definition}
        $\var{root\_concept} = \func{get\_unary\_concept}(\var{root\_variable})$\;
        $\var{root\_variable\_elements} = \func{get\_elements}(\var{root\_concept})$ \;
        
        \texttt{\\}
        \tcc{if no constituent elements (i.e. primitive concept), then create a point, set its properties and bind it.}
        \If{\textup{root\_variable\_elements.is\_empty}()}{
            \tcc{function defined below}
            $\func{generate\_primitive\_conjunctive\_unary\_relation}(\var{c}, \var{bindings})$\;
            \Return{\upshape{$\var{bindings}$}}}
         \texttt{\\}
        \tcc{composite concept: add binding of the root variable to its elements in the binding map}
        $\var{bindings}[\var{root\_variable}] = \var{root\_variable\_elements}$ \;
        \texttt{\\}
        \tcc{create a queue for the BFS and add the root variable to it}
        $\var{q} = \func{Queue}()$\;
        $\var{q}.\func{push}(\var{root\_variable})$ \;
         \texttt{\\}
        \tcc{perform BFS on the graph starting from the root variable.\;
            Mark nodes to avoid revisiting. Fail if there is an inconsistency}
        $\var{visited} = \func{Set}()$ \;
        
        \While{\upshape{$\var{q}$}}{
            \tcc{get next variable}
            $\var{v} = \var{q}.\func{pop}()$ \;
            \texttt{\\}
            \tcc{mark it as visited}
            $\var{visited}.\func{add}(\var{v})$\;
            \texttt{\\}
            \tcc{get concept for current variable}
            $\var{concept} = \func{get\_unary\_concept}(\var{v})$\;
            \texttt{\\}
            \tcc{generate the concept definition (recursively)}
            $\func{generate\_conjunctive}(\var{concept}, \var{bindings})$\;
             \texttt{\\}
            \tcc{generate any binary concepts r}
            \ForAll{\upshape{$\var{edges}$} ($\var{v}, \var{w}, \var{binary\_concept}$) $\in$ \textup{g} $\wedge  \var{w}$ \upshape{$\notin$} $\var{visited}$}
            {
             $\var{binary\_concept\_definition} = \func{get\_definition}(\var{binary\_concept})$\;
             $\func{generate\_binary\_concept}(\var{binary\_concept}, \var{bindings})$\;
             $\var{q}.\func{push}(\var{w})$
            }
            
            }
        
        \Return{\upshape{$\var{bindings}$}} \;
        \caption{\texttt{generate\_conjunctive\_concept}}
    \end{algorithm}
    \caption{Generation Algorithm for a given conjunctive concept.}
    \label{alg:conj-alg-gen}
\end{figure}

\begin{figure}[!ht]
    \begin{algorithm}[H]
    \KwIn{c : PrimitiveConjunctiveClause, bindings : Map}
    \tcc{generates a point object with properties determined by the supplied concept (e.g. $\texttt{x} \equiv \texttt{red}(x) \wedge x.x\_loc == 0 \wedge x.y\_loc == 16)$ }
    $\texttt{p} = \texttt{Point}()$ \;
    $\texttt{bindings}[\texttt{x}] = p$ \;
    \ForAll{\textup{unary} $\in$ \textup{c}}{
        \tcc{each primitive unary concept has its own generator implementation which sets the properties of the point.}
        \textup{generate\_primitive\_unary\_relation}(\textup{unary}, \textup{bindings})
    }
    \caption{\texttt{generate\_primitive\_conjunctive\_unary\_relation}}
    \end{algorithm}

    \begin{algorithm}[H]
    \SetKwProg{try}{try}{}{}
    \SetKwProg{catch}{catch}{}{end}
    \DontPrintSemicolon
    \SetNoFillComment
    \tcc{a binary concept $\texttt{r}(\mathbf{x},\mathbf{y})$ must be defined as a primitive relation using a reduction operator\;
    on both $\mathbf{x}$ and $\mathbf{y}$ to reduce them to points.\;
    for now these reduction operators are limited to argmin and argmax.\;
    Example: see $\texttt{east}$. There, the first reduction operator is $\argmax{}$; the second is $\argmin{}$\;
    and the primitive relation on them is $\texttt{east\_point}$} \;
    $\texttt{reduction}_1 = \texttt{get\_reduction}_1(\texttt{c})$ \;
    $\texttt{reduction}_2 = \texttt{get\_reduction}_2(\texttt{c})$ \;
    $\texttt{prim\_concept} = \texttt{get\_prim\_concept}(\texttt{c})$ \;
    $x_1 = \texttt{reduction}_1(\mathbf{x_1})$ \;
    \ForAll{\upshape{pairs} ($\var{x}_1$, $\var{y}$), $\var{y} \in \mathbf{y}$}{
      \try{} {
          \tcc{the primitive generators must be defined individually for each primitive concept}
          $\texttt{generate\_binary\_concept\_primitive}$($\texttt{prim\_concept}$, $x_1$, $y$, $\var{bindings})$ \;
          \tcc{check that the reduction over y is satisfied by the generated bindings}
          $\func{check\_reduction}$(\var{$\mathbf{y}$}, $\var{reduction}_2$, $\var{bindings})$ \;
        }
        \catch {} {
          $\texttt{continue}$
        }
    } \;
    
    \caption{\texttt{generate\_binary\_concept}}
    \end{algorithm}
    \caption{Specific functions, used in Alg.~\ref{alg:conj-alg-gen}.}
    \label{alg:specific-functions}
\end{figure}

\clearpage
\subsection{Additional concept examples}

\begin{figure}[ht]
    \centering
    \includegraphics[width=\textwidth]{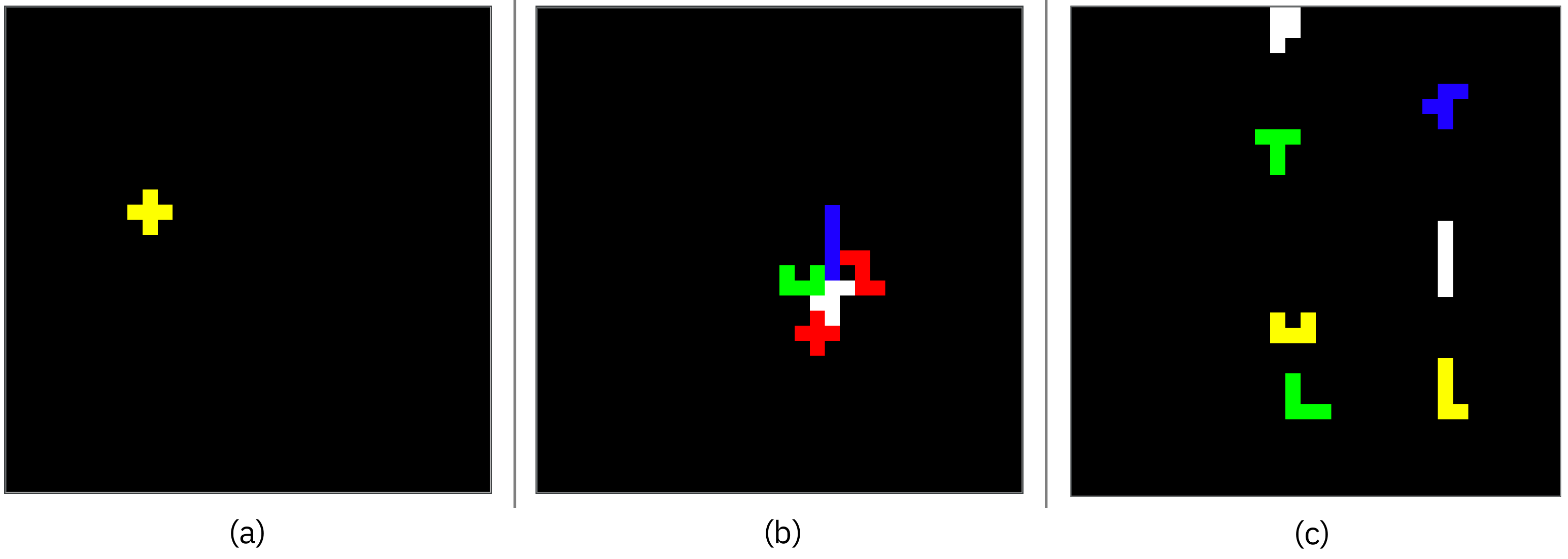}
    \caption{Some examples of additional concepts not included in our experiments. (a): the "X" pentomino. (b): An "X" pentomino shape ("+") made of "I", "U", "F", "Z", "X" pentominoes. (c): two loose stacks of pentominoes, one east of the other.}
    \label{fig:additional_samples}
\end{figure}

Fig.~\ref{fig:additional_samples} presents some examples of concepts which we generated with ConceptWorld but did not include in our experiments.

\clearpage
\section{Experiment 1: Complete set of results and additional details}
\label{app:exp_1}

\subsection{Training details and hyper-parameters search}
\label{app:training_details}
We developed the codebase using PyTorch~\cite{pytorch}. All experiments being multi-class image classification tasks, we used Adam~\cite{kingma2014adam} and Cross-Entropy loss. We fixed the number of epochs to 100, and early-stopped, based on the validation loss. The default learning rate was $1^{-3}$ and the batch size 128. We used different GPU-accelerated platforms, with NVIDIA’s GeForce GTX TITAN X and Tesla K80 GPUs.

For the CNN, MLP and ResNet~\cite{he2016deep}, we started with published or common hyper-parameter values:
\begin{itemize}
    \item CNN: 2 convolution layers, each with output size 128, kernel size 5, stride 1, padding 0; 2 maxpool layers of kernel size 2.
    \item MLP: 1 hidden layer of size 128, along with input and output layers.
    \item ResNet: we used ResNet18, for which we replaced the last 2 layers (layer 3 and 4) by identity functions. This was done to reduce the number of parameters, making it comparable to the other baselines.
\end{itemize}

For the 3 models above, we performed random hyper-parameter search, using Experiment 1 (pure and mixed sequences of 2x2 squares) over the learning rate, hidden size and number of layers, optimizing for validation accuracy. We found that the original hyper-parameters worked well for all experiments, and therefore used them for all reported experiments.

In contrast, we observed that WReN and PrediNet are sensitive to hyper-parameters. Starting with the published values, we performed random search over Experiment 1  -- verifying that performance was equally good in other experiments -- over the learning rate, number and size of the input convolutional layers, as well as the key size in the case of PrediNet. 

We obtained, selecting based on validation accuracy, the following sets of hyperparameters for PrediNet and WReN:

PrediNet:
\begin{itemize}
    \item Learning rate: $1^{-5}$,
    \item Training batch size: 64,
    \item Key size: 16,
    \item Number of attention heads: 2,
    \item Number of relations: 2,
    \item 1 input convolutional layer (with bias and batch normalization) with output size 32, kernel size 12, stride 6.
\end{itemize}

For WReN:
\begin{itemize}
    \item Learning rate: $1^{-5}$,
    \item Training batch size: 128,
    \item Key size: 16,
    \item Number of attention heads: 2,
    \item 3 input convolutional layers (with batch normalization) with output size 64, kernel size 2, stride 2.
\end{itemize}

With these parameter values, all models have comparable size, between 400k and 600k trainable parameters.

\subsection{Additional examples}

\begin{figure}[ht]
    \centering
    \includegraphics[width=0.7\textwidth]{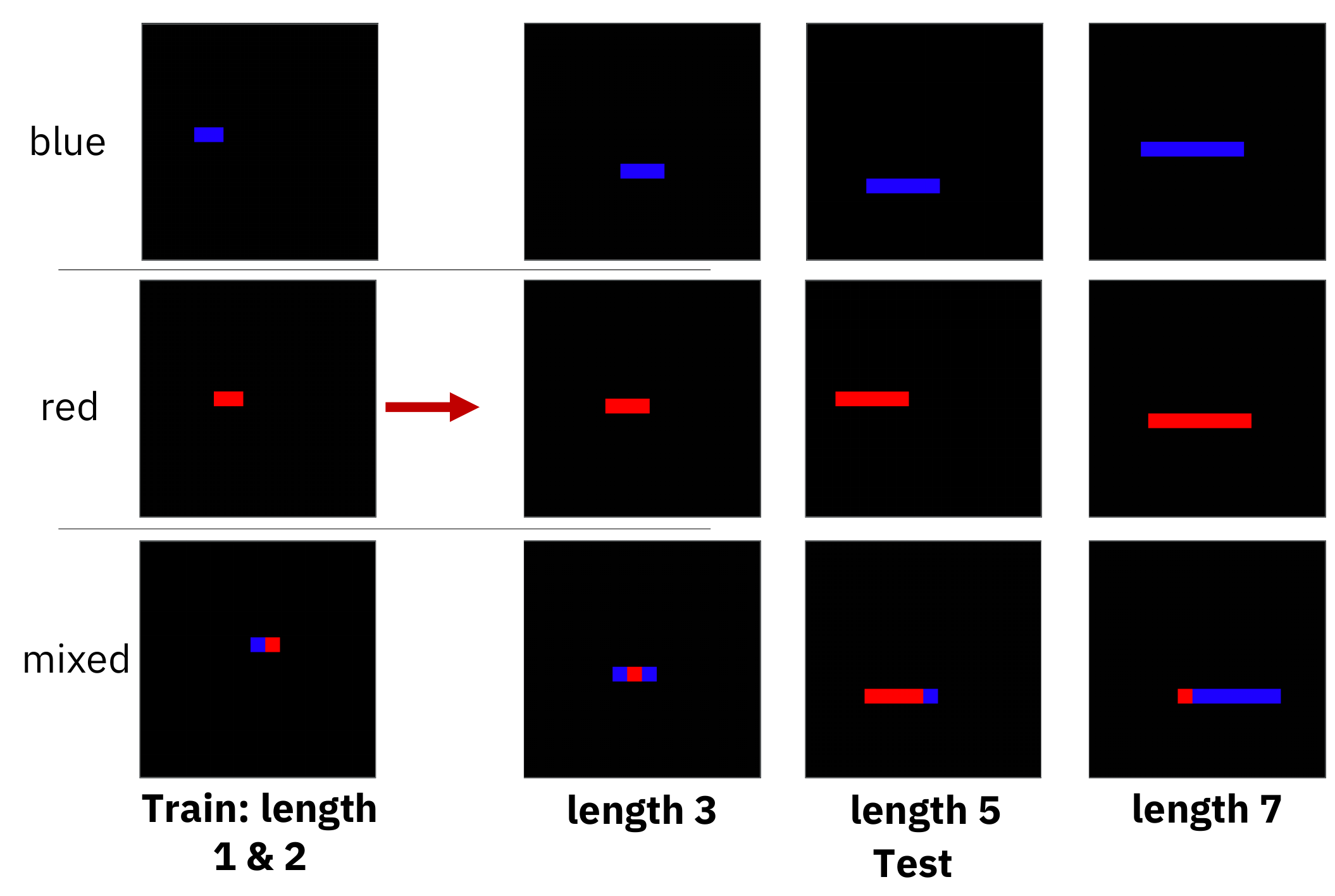}
    \caption{Some examples of the \textit{all-red}, \textit{all-blue} and \textit{mixed} sequences considered in Experiment 1, for the training and test sets (\textit{best viewed in color}).}
    \label{fig:exp_1_samples}
\end{figure}

Fig.~\ref{fig:exp_1_samples} contains some examples of the sequences we considered in this experiment.

\subsection{Results}

The number of samples per concept for the training and test sets is available in Table~\ref{tab:nb_samples_exp_1}. The training set is an union over sequences of length 1 and 2, except for the \textit{mixed} sequences, which can only be defined from length 2 and up. The test sets, one per sequence length, are independent from one another.

\begin{table}[!ht]
	\centering
	\begin{tabular}{cccccc}
		\toprule
		\multirow{2}{*}{Concept} & \multicolumn{2}{c}{Train} & \multicolumn{3}{c}{Test} \\
		\cmidrule(lr){2-3} \cmidrule(lr){4-6}
		 & Seq len 1 & Seq len 2 & Seq len 3 & Seq len 5 & Seq len 7 \\
		 \midrule
		 red & 800 & 800 & 600 & 600 & 500 \\
		\midrule
		 blue & 800 & 800 & 600 & 600 & 500 \\
		 \midrule
		 mixed & - & 1600 & 600 & 600 & 500 \\
		\bottomrule
	\end{tabular}
	\smallskip
	\caption{Number of samples per concept and sequence length for Experiment 1. Due to the constant image size (32 x 32), the longer the sequence, the smaller the space of corresponding samples.}
	\label{tab:nb_samples_exp_1}
\end{table}

Table \ref{tab:exp_1_results_full} shows F1 scores per class, sorted by model and sequence length. These scores were obtained by averaging over 10 runs, each with a different random seed.

\begin{table}[!ht]
	\centering
	\begin{tabular}{cccccc}
		\toprule
		\multicolumn{2}{l}{\multirow{2}{*}{Model}} & \multirow{2}{*}{Concept} & \multicolumn{3}{c}{F1 score on test set} \\
		\cmidrule{4-6}
		\multicolumn{2}{l}{} & & Seq len 3 & Seq len 5 & Seq len 7 \\
		\midrule
		\multirow{6}{*}{ResNet} & \multirow{3}{*}{Pretrained} & Blue & 0.993 & 0.935 & 0.890 \\
		\cmidrule{3-6} 
		\multicolumn{2}{l}{} & Red & 0.997 & 0.929 & 0.881 \\
		\cmidrule{3-6}
		\multicolumn{2}{l}{} & Mixed & 0.990 & 0.827 & 0.646 \\
		\cmidrule{2-6}
		 & \multirow{3}{*}{Non-Pretrained} & Blue & 0.991 & 0.913 & 0.867 \\
		\cmidrule{3-6}
		\multicolumn{2}{l}{} & Red & 0.992 & 0.905 & 0.855 \\
		\cmidrule{3-6}
		\multicolumn{2}{l}{} & Mixed & 0.981 & 0.742 & 0.512 \\
		\midrule 
		\multicolumn{2}{l}{\multirow{3}{*}{WReN}} & Blue & 0.921 & 0.842 & 0.814 \\
		\cmidrule{3-6}
		\multicolumn{2}{l}{} & Red & 0.915 & 0.848 & 0.809 \\
		\cmidrule{3-6}
		\multicolumn{2}{l}{} & Mixed & 0.828 & 0.499 & 0.350 \\
		\midrule
		\multicolumn{2}{l}{\multirow{3}{*}{SimpleConvNet}} & Blue & 0.964 & 0.829 & 0.803 \\
		\cmidrule{3-6}
		\multicolumn{2}{l}{} & Red & 0.945 & 0.844 & 0.822 \\
		\cmidrule{3-6}
		\multicolumn{2}{l}{} & Mixed & 0.929 & 0.304 & 0.05 \\
		\midrule
		\multicolumn{2}{l}{\multirow{3}{*}{PrediNet}} & Blue & 0.795 & 0.709 & 0.689 \\
		\cmidrule{3-6}
		\multicolumn{2}{l}{} & Red & 0.798 & 0.733 & 0.717 \\
		\cmidrule{3-6}
		\multicolumn{2}{l}{} & Mixed & 0.705 & 0.321 &0.195 \\
		\midrule
		\multicolumn{2}{l}{\multirow{3}{*}{SimpleFeedForward}} & Blue & 0.805 & 0.800 & 0.800 \\
		\cmidrule{3-6}
		\multicolumn{2}{l}{} & Red & 0.818 & 0.800 & 0.800 \\
		\cmidrule{3-6}
		\multicolumn{2}{l}{}& Mixed & 0.119 & 0 & 0 \\
		\bottomrule
	\end{tabular}
	\smallskip
	\caption{F1 scores per concept and test sequence length for all 5 baselines in Experiment 1.}
	\label{tab:exp_1_results_full}
\end{table}

\clearpage
\section{Experiment 2: Complete set of results and additional details}
\label{app:exp_2}

\begin{figure}[ht]
    \centering
    \includegraphics[width=\textwidth]{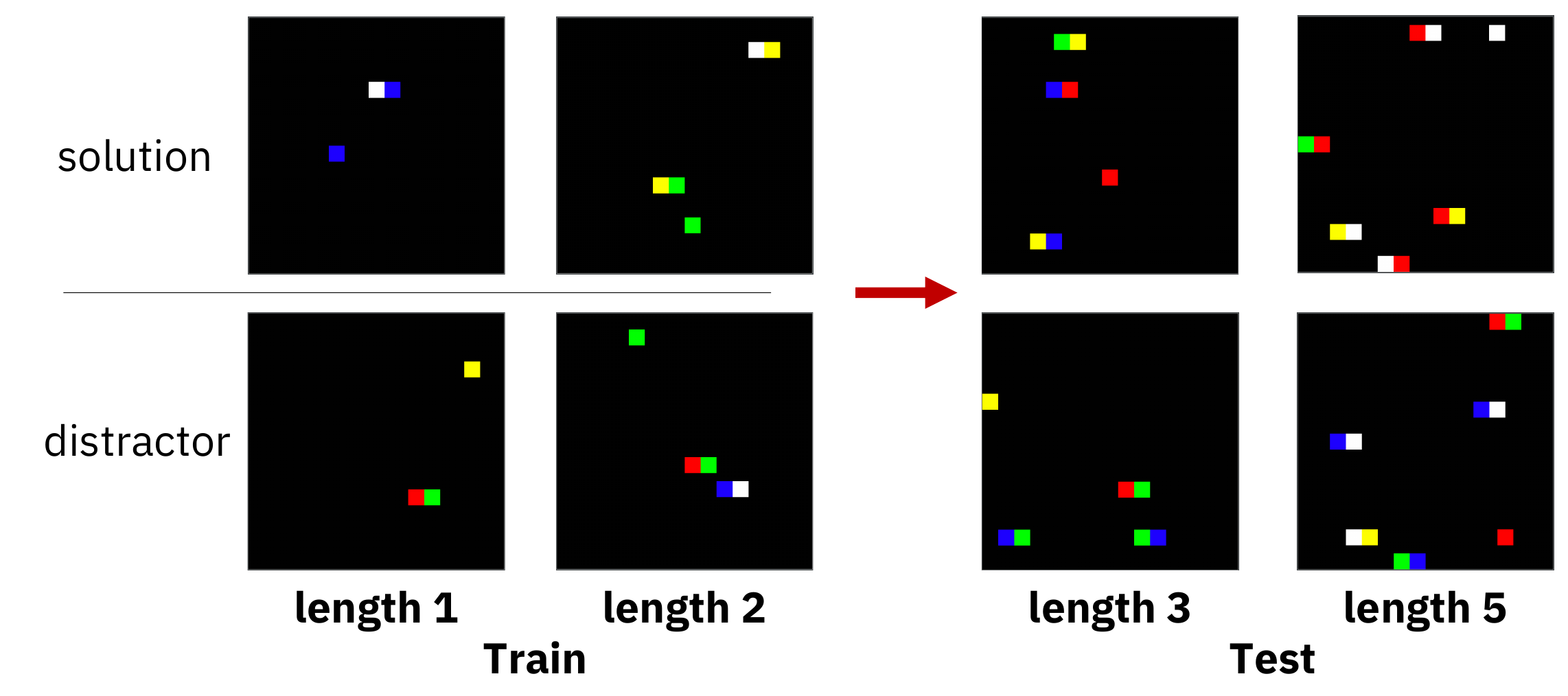}
    \caption{Some examples of the \texttt{solution} and \texttt{distractor} concepts in Experiment 2 (\textit{best viewed in color}).}
    \label{fig:exp_2_samples}
\end{figure}

Fig.~\ref{fig:exp_2_samples} illustrates some of the valid and invalid $key-lock$ paths of different lengths we considered for Experiment 2 (Box-World).

The number of samples per concept for the training and test sets is available in Table~\ref{tab:nb_samples_exp_2}. The training set is an union over paths of length 1 (1 $key-lock$ pair) and 2 (2 $key-lock$ pairs). The test sets, one per sequence length, are independent from one another.

\begin{table}[!ht]
	\centering
	\begin{tabular}{ccccc}
		\toprule
		\multirow{2}{*}{Concept} & \multicolumn{2}{c}{Train} & \multicolumn{2}{c}{Test} \\
		\cmidrule(lr){2-3} \cmidrule(lr){4-5}
		 & Seq len 1 & Seq len 2 & Seq len 3 & Seq len 5 \\
		 \midrule
		 solution & 500 & 500 & 500 & 500 \\
		 \midrule
		 distractor & 500 & 500 & 500 & 500 \\
		 \midrule
		 2x2 square & \multicolumn{2}{c}{1000} & \multicolumn{2}{c}{500} \\
		\bottomrule
	\end{tabular}
	\smallskip
	\caption{Number of samples per concept and sequence length for Experiment 2. The \texttt{2x2 square} concept is independent of sequence length, and the same samples are used for both test sets.}
	\label{tab:nb_samples_exp_2}
\end{table}

Table~\ref{tab:exp_2_results_full} contains the F1 scores over the test sets, for the 5 baselines. These scores were also obtained by averaging over 10 runs.

\begin{table}[!ht]
	\centering
	\begin{tabular}{cccc}
		\toprule
		\multirow{2}{*}{Model} & \multirow{2}{*}{Concept} & \multicolumn{2}{c}{F1 score on test set} \\
		\cmidrule{3-4}
		& & Seq len 3 & Seq len 5 \\
		\midrule
		\multirow{3}{*}{ResNet (Pretrained)} & Solution & 0.301 & 0.303 \\
		\cmidrule{2-4} 
		& Distractor & 0.427 & 0.411 \\
		\cmidrule{2-4} 
		& 2x2 Square & 0.999 & 0.999 \\
		\midrule 
		\multirow{3}{*}{WReN} & Solution & 0.584 & 0.642 \\
		\cmidrule{2-4} 
		& Distractor & 0.379 & 0.154 \\
		\cmidrule{2-4} 
		& 2x2 Square & 0.986 & 0.986 \\
		\midrule 
		\multirow{3}{*}{SimpleConvNet} & Solution & 0.465 & 0.447 \\
		\cmidrule{2-4} 
		& Distractor & 0.439 & 0.412 \\
		\cmidrule{2-4} 
		& 2x2 Square & 1 & 1 \\
		\midrule 
		\multirow{3}{*}{PrediNet} & Solution & 0.361 & 0.368 \\
		\cmidrule{2-4} 
		& Distractor & 0.416 & 0.398 \\
		\cmidrule{2-4} 
		& 2x2 Square & 0.551 & 0.551 \\
		\midrule 
		\multirow{3}{*}{SimpleFeedForward} & Solution & 0.447 & 0.441 \\
		\cmidrule{2-4} 
		& Distractor & 0.456 & 0.447 \\
		\cmidrule{2-4} 
		& 2x2 Square & 0.901 & 0.901 \\
		\bottomrule
	\end{tabular}
	\smallskip
	\caption{F1 scores per concept and test sequence length for all 5 baselines in Experiment 2. The test samples for the \texttt{2x2 square} concept are the same for both sequence lengths.}
	\label{tab:exp_2_results_full}
\end{table}

\clearpage
\section{Experiment 3: Complete set of results and additional details}
\label{app:exp_3}

\begin{figure}[ht]
    \centering
    \includegraphics[width=0.8\textwidth]{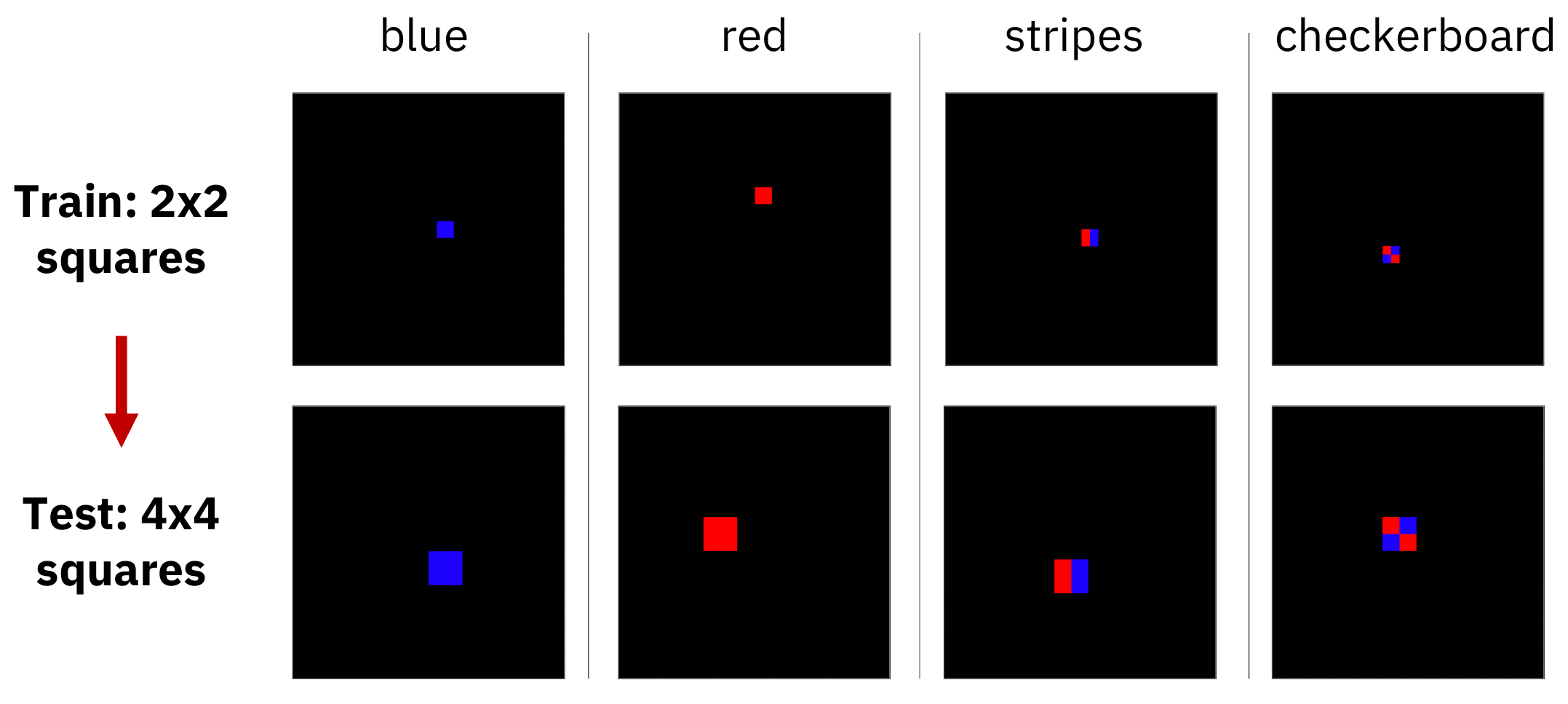}
    \caption{Some examples of the 4 concepts in Experiment 3 (\textit{best viewed in color}).}
    \label{fig:exp_3_samples}
\end{figure}

Fig.~\ref{fig:exp_3_samples} illustrates some of the train and test samples we created for the \textit{all blue}, \textit{all red}, \textit{vertical alternating red/blue stripes}, and \textit{checkerboard pattern of red/blue} concepts in Experiment 3.

Table~\ref{tab:nb_samples_exp_3} contains the number of samples for the training (2x2 squares) and test (4x4 squares) for each concept.

\begin{table}[!ht]
	\centering
	\begin{tabular}{ccc}
		\toprule
		Concept & Train (2x2 squares) & Test (4x4 squares) \\
		\midrule
		 all blue & 700 & 350 \\
		 \midrule
		 all red & 700 & 350 \\
		 \midrule
		 stripes & 800 & 350 \\
		 \midrule
		 checkerboard & 800 & 350 \\
		\bottomrule
	\end{tabular}
	\smallskip
	\caption{Number of samples per concept for Experiment 3.}
	\label{tab:nb_samples_exp_3}
\end{table}

Table \ref{tab:exp_3_results_full} shows the F1 score per concept and model.

\begin{table}[!ht]
	\centering
	\begin{tabular}{ccc}
		\toprule
		Model & Concept & F1 score on test set \\
		\midrule
		\multirow{4}{*}{ResNet (Pretrained)} & Blue & 0.7416 \\
		\cmidrule{2-3}
		& Red & 0.8295 \\
		\cmidrule{2-3}
		& Vertical Stripes & 0.3221 \\
		\cmidrule{2-3}
		& Checkerboard & 0.3807 \\
		\midrule 
		\multirow{4}{*}{WReN} & Blue & 0.4645 \\
		\cmidrule{2-3}
		& Red & 0.5293 \\
		\cmidrule{2-3}
		& Vertical Stripes & 0.10189 \\
		\cmidrule{2-3}
		& Checkerboard & 0.2415 \\
		\midrule 
		\multirow{4}{*}{SimpleConvNet} & Blue & 0.97879 \\
		\cmidrule{2-3}
		& Red & 0.984699 \\
		\cmidrule{2-3}
		& Vertical Stripes & 0.6614 \\
		\cmidrule{2-3}
		& Checkerboard & 0.3566 \\
		\midrule 
		\multirow{4}{*}{PrediNet} & Blue & 0.2388 \\
		\cmidrule{2-3}
		& Red & 0.3063 \\
		\cmidrule{2-3}
		& Vertical Stripes & 0.183 \\
		\cmidrule{2-3}
		& Checkerboard & 0.1833 \\
		\midrule 
		\multirow{4}{*}{SimpleFeedForward} & Blue & 0.9998 \\
		\cmidrule{2-3}
		& Red & 1 \\
		\cmidrule{2-3}
		& Vertical Stripes & 0.5157 \\
		\cmidrule{2-3}
		& Checkerboard & 0.5074 \\
		\bottomrule
	\end{tabular}
	\smallskip
	\caption{F1 scores per concept for all 5 baselines for Experiment 3.}
	\label{tab:exp_3_results_full}
\end{table}

\clearpage
\section{Experiment 4: Complete set of results and some examples}
\label{app:visually-different-substitute}

\begin{figure}[ht]
    \centering
    \includegraphics[width=\textwidth]{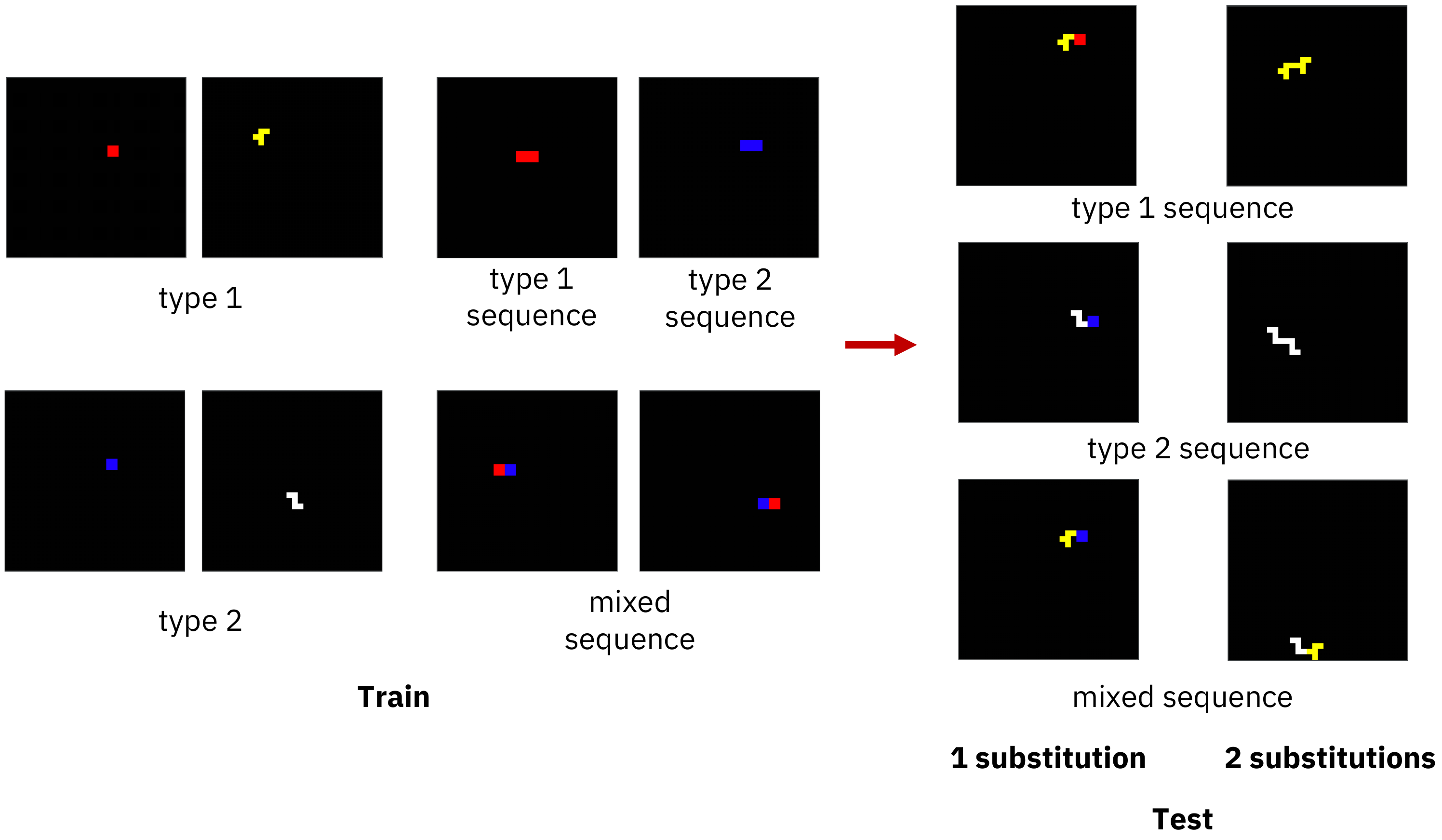}
    \caption{Some examples of the \texttt{type 1} and \texttt{type 2} classes, as well as the "pure" and mixed sequences built from both types (\textit{best viewed in color}).}
    \label{fig:exp_4_samples}
\end{figure}

Fig.~\ref{fig:exp_4_samples} showcases the \texttt{type 1} and \texttt{type 2} classes, as well as the pairs we built with them. Test pairs involve 1 or 2 substitutions.

The details of the training and test sets are available in Table~\ref{tab:nb_samples_exp_4}. The \texttt{type 1} and \texttt{type 2} classes are an union over 2 concepts (i.e. \textit{2x2 red square} and \textit{"F" pentomino} for \texttt{type 1}). The number of samples for these 2 classes is thus equidistributed between the 2 concepts.

\begin{table}[!ht]
	\centering
	\begin{tabular}{cccc}
		\toprule
		\multirow{2}{*}{Concept} & \multirow{2}{*}{Train} & \multicolumn{2}{c}{Test} \\
		\cmidrule(lr){3-4}
		 & & 1 substitution & 2 substitutions \\
		 \midrule
		 type 1 & 400 + 400 & \multicolumn{2}{c}{200 + 200} \\
		 \midrule
		 type 2 & 400 + 400 & \multicolumn{2}{c}{200 + 200} \\
		 \midrule
		 type 1 sequences & 800 & 400 & 400 \\
		 \midrule
		 type 2 sequences & 800 & 400 & 400 \\
		 \midrule
		 mixed sequences & 800 & \multicolumn{2}{c}{400} \\
		\bottomrule
	\end{tabular}
	\smallskip
	\caption{Number of samples per concept and number of substitutions for Experiment 4. The \texttt{type 1} and \texttt{type 2} concepts are independent of the number of substitutions: the same samples are used for both tests.}
	\label{tab:nb_samples_exp_4}
\end{table}

Table \ref{tab:exp_4_results_full} contains the F1 scores per concept, for both test sets, and all baselines.

\begin{table}[!ht]
	\centering
	\begin{tabular}{cccc}
		\toprule
		\multirow{2}{*}{Model} & \multirow{2}{*}{Concept} & \multicolumn{2}{c}{F1 score on test set} \\
		\cmidrule{3-4}
		& & 1 substitution & 2 substitutions \\
		\midrule
		\multirow{5}{*}{ResNet (Pretrained)} & \texttt{Type 1} & 0.603 & 0.6008 \\
		\cmidrule{2-4} 
		& \texttt{Type 2} & 0.5774 & 0.5479 \\
		\cmidrule{2-4} 
		& \texttt{Type 1} Sequences & 0.0517 & 0 \\
		\cmidrule{2-4} 
		& \texttt{Type 2} Sequences & 0.0941 & 0.0005 \\
		\cmidrule{2-4} 
		& Mixed Sequences & 0.1159 & 0.001 \\
		\midrule 
		\multirow{5}{*}{WReN} & \texttt{Type 1} & 0.582 & 0.5246 \\
		\cmidrule{2-4} 
		& \texttt{Type 2} & 0.594 & 0.5486 \\
		\cmidrule{2-4} 
		& \texttt{Type 1} Sequences & 0.2974 & 0.0324 \\
		\cmidrule{2-4} 
		& \texttt{Type 2} Sequences & 0.2031 & 0.0095 \\
		\cmidrule{2-4} 
		& Mixed Sequences & 0.207 & 0.0457 \\
		\midrule 
		\multirow{5}{*}{SimpleConvNet} & \texttt{Type 1} & 0.5806 & 0.5612 \\
		\cmidrule{2-4} 
		& \texttt{Type 2} & 0.586 & 0.581 \\
		\cmidrule{2-4} 
		& \texttt{Type 1} Sequences & 0.0908 & 0 \\
		\cmidrule{2-4} 
		& \texttt{Type 2} Sequences & 0.0603 & 0 \\
		\cmidrule{2-4} 
		& Mixed Sequences & 0.1157 & 0 \\
		\midrule 
		\multirow{5}{*}{PrediNet} & \texttt{Type 1} & 0.3656 & 0.3324 \\
		\cmidrule{2-4} 
		& \texttt{Type 2} & 0.4705 & 0.4235 \\
		\cmidrule{2-4} 
		& \texttt{Type 1} Sequences & 0.2854 & 0.0759 \\
		\cmidrule{2-4} 
		& \texttt{Type 2} Sequences & 0.357 & 0.0997 \\
		\cmidrule{2-4} 
		& Mixed Sequences & 0.2023 & 0.081 \\
		\midrule 
		\multirow{5}{*}{SimpleFeedForward} & \texttt{Type 1} & 0.3916 & 0.3291 \\
		\cmidrule{2-4} 
		& \texttt{Type 2} & 0.4083 & 0.3989 \\
		\cmidrule{2-4} 
		& \texttt{Type 1} Sequences & 0.217 & 0.0039 \\
		\cmidrule{2-4} 
		& \texttt{Type 2} Sequences & 0.0748 & 0 \\
		\cmidrule{2-4} 
		& Mixed Sequences & 0.1757 & 0.0326 \\
		\bottomrule
	\end{tabular}
	\smallskip
	\caption{F1 scores per concept for Experiment 4.}
	\label{tab:exp_4_results_full}
\end{table}

\subsection{Curriculum Training}
To further study the potential impact of learning the \texttt{type 1} and \texttt{type 2} concepts concurrently with the higher-order concepts, we tested a curriculum variant. Here, we trained the models until convergence (20 epochs) on \texttt{type 1} and \texttt{type 2} ($1^{st}$ curriculum stage), then added the \texttt{type 1}, \texttt{type 2} and \texttt{mixed} sequences ($2^{nd}$ curriculum stage). Table~\ref{tab:exp_4_results_curr} contains the F1 scores per concept for each model. The trained models were selected using the validation loss during the $2^{nd}$ curriculum stage. For most models, curriculum training resulted in worse F1 scores compared to non-curriculum training.

\begin{table}[!ht]
	\centering
	\begin{tabular}{cccc}
		\toprule
		\multirow{2}{*}{Model} & \multirow{2}{*}{Concept} & \multicolumn{2}{c}{F1 score on test set} \\
		\cmidrule{3-4}
		& & 1 substitution & 2 substitutions \\
		\midrule
		\multirow{5}{*}{ResNet (Pretrained)} & \texttt{Type 1} & 0.6857	& 0.5919 \\
		\cmidrule{2-4} 
		& \texttt{Type 2} & 0.7071 & 0.5721 \\
		\cmidrule{2-4} 
		& \texttt{Type 1} Sequences & 0.2522 & 0.0594 \\
		\cmidrule{2-4} 
		& \texttt{Type 2} Sequences & 0.3024 & 0.0067 \\
		\cmidrule{2-4} 
		& Mixed Sequences & 0.3785 & 0.244 \\
		\midrule 
		\multirow{5}{*}{WReN} & \texttt{Type 1} & 0.4751 & 0.4365 \\
		\cmidrule{2-4} 
		& \texttt{Type 2} & 0.4327 & 0.4038 \\
		\cmidrule{2-4} 
		& \texttt{Type 1} Sequences & 0.2935 & 0.0892 \\
		\cmidrule{2-4} 
		& \texttt{Type 2} Sequences & 0.197 & 0.0224 \\
		\cmidrule{2-4} 
		& Mixed Sequences & 0.2415 & 0.1642 \\
		\midrule 
		\multirow{5}{*}{SimpleConvNet} & \texttt{Type 1} & 0.5576 & 0.5559 \\
		\cmidrule{2-4} 
		& \texttt{Type 2} & 0.5861 & 0.5826 \\
		\cmidrule{2-4} 
		& \texttt{Type 1} Sequences & 0.001 & 0 \\
		\cmidrule{2-4} 
		& \texttt{Type 2} Sequences & 0.005 & 0 \\
		\cmidrule{2-4} 
		& Mixed Sequences & 0.0541 & 0 \\
		\midrule 
		\multirow{5}{*}{PrediNet} & \texttt{Type 1} & 0.2393 & 0.235 \\
		\cmidrule{2-4} 
		& \texttt{Type 2} & 0.1333 & 0.1233 \\
		\cmidrule{2-4} 
		& \texttt{Type 1} Sequences & 0.1834 & 0.1179 \\
		\cmidrule{2-4} 
		& \texttt{Type 2} Sequences & 0.2958 & 0.232 \\
		\cmidrule{2-4} 
		& Mixed Sequences & 0.138 & 0.1453 \\
		\midrule 
		\multirow{5}{*}{SimpleFeedForward} & \texttt{Type 1} & 0.292 & 0.2965 \\
		\cmidrule{2-4} 
		& \texttt{Type 2} & 0.258 & 0.2476 \\
		\cmidrule{2-4} 
		& \texttt{Type 1} Sequences & 0.049 & 0.0274 \\
		\cmidrule{2-4} 
		& \texttt{Type 2} Sequences & 0.0155 & 0.0103 \\
		\cmidrule{2-4} 
		& Mixed Sequences & 0.0253 & 0.0158 \\
		\bottomrule
	\end{tabular}
	\smallskip
	\caption{F1 scores per concept for the curriculum variant of Experiment 4.}
	\label{tab:exp_4_results_curr}
\end{table}

\end{document}